\documentclass{article}
\usepackage{array}

\usepackage{microtype}
\usepackage{graphicx}

\usepackage{tocloft}

\usepackage{amsmath}

\usepackage{hyperref}
\usepackage[accepted]{icml2024}

\usepackage{color, colortbl}
\usepackage{bbm}
\usepackage{amsfonts}       %
\usepackage{multirow}
\usepackage{pifont}
\usepackage{enumerate}
\usepackage[T1]{fontenc}

\usepackage{caption}
\usepackage{subcaption}
\usepackage[toc,page,header]{appendix}

\newtheorem{theorem}{Theorem}[section]

\newtheorem{definition}[theorem]{Definition}

\newcommand{\bcmark}{\textcolor{black}{\ding{51}}} %
\newcommand{\bxmark}{\textcolor{black}{\ding{55}}} %
\newcommand{\eg}{\textit{e}.\textit{g}.}
\newcommand{\ie}{\textit{i}.\textit{e}.}
\newcommand{\etal}{et al.}
\definecolor{Gray}{gray}{0.9}

\newcommand{\expnum}[2]{{#1}\mathrm{e}{-#2}}

\usepackage[textsize=tiny]{todonotes}

\icmltitlerunning{Improving Robustness to Multiple Spurious Correlations by Multi-Objective Optimization}

\begin{document}

\twocolumn[
\icmltitle{
Improving Robustness to Multiple Spurious Correlations\\by Multi-Objective Optimization
}
\begin{icmlauthorlist}
\icmlauthor{Nayeong Kim}{yyy}
\icmlauthor{Juwon Kang}{yyy}
\icmlauthor{Sungsoo Ahn}{yyy,comp}
\icmlauthor{Jungseul Ok}{yyy,comp}
\icmlauthor{Suha Kwak}{yyy,comp}
\end{icmlauthorlist}

\icmlaffiliation{yyy}{Department of Computer Science and Engineering, POSTECH, Pohang, Korea}
\icmlaffiliation{comp}{Graduate School of Artificial Intelligence, POSTECH, Pohang, Korea}

\icmlcorrespondingauthor{Suha Kwak}{suha.kwak@postech.ac.kr}

\icmlkeywords{Machine Learning, ICML}

\vskip 0.3in
]

\printAffiliationsAndNotice{}  %

\begin{abstract}
We study the problem of training an unbiased and accurate model given a dataset with multiple biases. This problem is challenging since the multiple biases cause multiple undesirable shortcuts during training, and even worse, mitigating one may exacerbate the other.
We propose a novel training method to tackle this challenge.
Our method first groups training data so that different groups induce different shortcuts, and then optimizes a linear combination of group-wise losses while adjusting their weights dynamically to alleviate conflicts between the groups in performance; this approach, rooted in the multi-objective optimization theory, encourages to achieve the minimax Pareto solution.
We also present a new benchmark with multiple biases, dubbed MultiCelebA, for evaluating debiased training methods under realistic and challenging scenarios. Our method achieved the best on three datasets with multiple biases, and also showed superior performance on conventional single-bias datasets.

\end{abstract}

\section{Introduction}
\begin{figure}[h!]
\centering
\includegraphics[width=\linewidth]{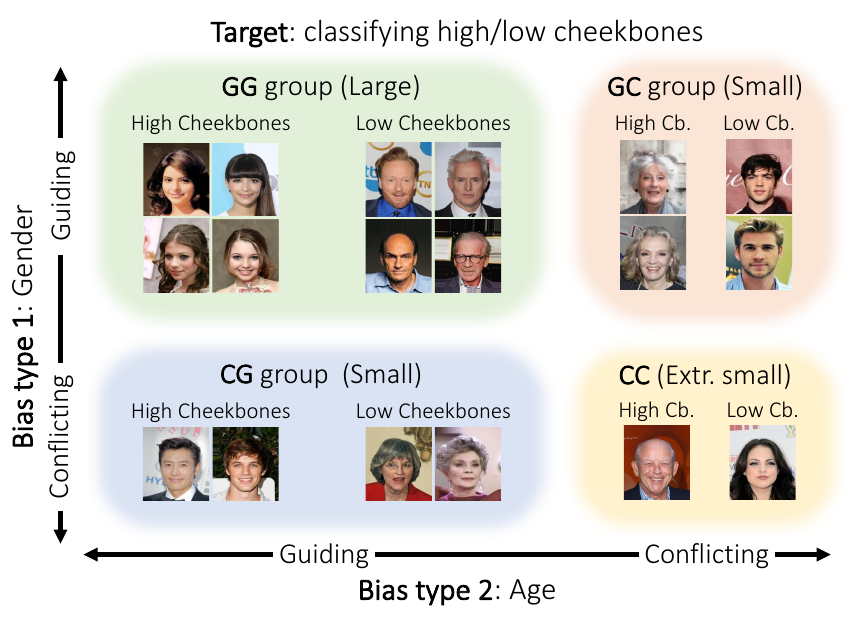}
\caption{
A example of grouping training data with two bias types. Each axis represents each bias type, for which bias-guiding samples make up the majority and bias-conflicting ones hold the minority.
In this example, the name of each group indicates if samples of the group has a guiding attribute (\textsc{G}) or a conflicting attribute (\textsc{C}) for \texttt{gender} and \texttt{age}, in respective order.
}
\vspace{-2mm}
\label{fig:grouping}
\end{figure} 

Empirical risk minimization (ERM)~\citep{vapnik1999nature} is currently the gold standard 
in supervised learning of deep neural networks.
However, recent studies \citep{sagawa2019distributionally,geirhos2020shortcut} revealed 
that ERM is prone to taking \textit{undesirable shortcuts} stemming from \textit{spurious correlations} between the target labels and irrelevant attributes arising from subgroup imbalance of training data.
For example, \citet{sagawa2019distributionally} showed how much a deep neural network trained to classify bird species relies on the background rather than the bird itself.
Such a spurious correlation is often hard to mitigate since the data collection procedure itself is biased towards the correlation.

To resolve this issue, researchers have studied debiased training algorithm, \ie, algorithms training a model while mitigating spurious correlations \citep{arjovsky2019invariant,bahng2020learning,sagawa2019distributionally,teney2020unshuffling,tartaglione2021end, LDR, LfF, JTT, kim2022learning}. 
They focus on improving performance on bias-conflicting samples (\ie, samples that disagree with the spurious correlations) to achieve a balance of bias-conflicting and bias-guiding samples (\ie, those agreeing with the spurious correlations).
Although these algorithms have shown promising results, they have been evaluated in a limited setting where only a single type of spurious correlation exists in training data.

We advocate that debiased training algorithms should be evaluated under more realistic scenarios with multiple biases.
In such scenarios, some samples may align with one bias but may conflict with another, which makes mitigating spurious correlations more challenging. 
If one only considers the intersection of bias-conflicting samples, \ie, \textit{clean} samples that disagree with all the spurious correlations, the resulting group will be extremely small as illustrated in Figure~\ref{fig:grouping} and result in overfitting consequently. 
Furthermore, mitigation of one bias often promote another as empirically observed by \citet{li2023whac}.
These complexities often lead existing methods to achieve similar or even worse performance compared to simple baselines such as upsampling and upweighting. %

\begin{figure}[t!]
\centering
\includegraphics[width=\linewidth]{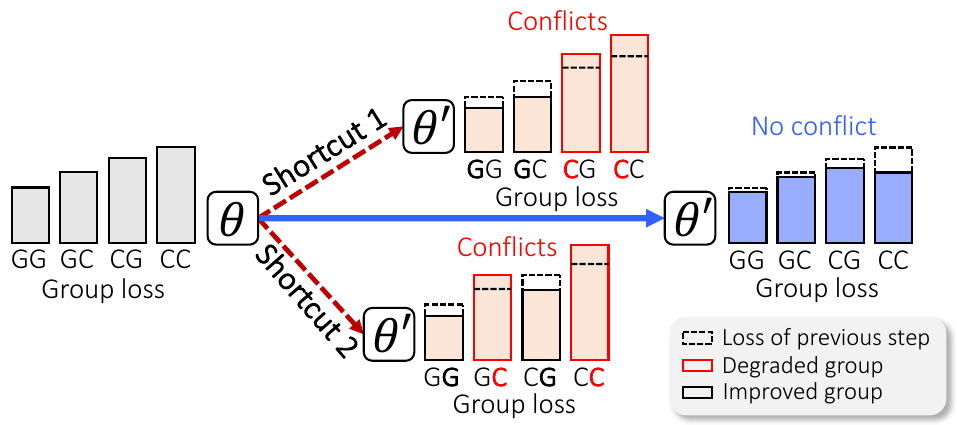}
\caption{
The concept of between-group conflicts and model biases.
During model parameter updates ($\theta$), the model risks exploiting spurious correlations as shortcuts for classification (red lines). 
Updating model parameters toward shortcut results in a reduced group-wise loss in the guiding groups but amplifies the loss in the conflicting groups (\eg, \textsc{CG} and \textsc{CC} for shortcut 1), leading to group conflicts. 
Updating parameters towards cues directly related to the target classification, free from spurious correlations (blue line), offers the only solution to minimize losses across all groups.
}
\vspace{-2mm}
\label{fig:idea}
\end{figure}

We propose a novel debiased training algorithm to tackle the aforementioned challenges.
Our algorithm first divides the entire training set into multiple groups where data of the same group have the same impact on training in terms of the model bias, \ie, guiding to or conflicting with each bias type in the same way, as illustrated in Figure~\ref{fig:grouping}.
With the grouping strategy, all groups share the same target task (\ie, classifying cheekbones), but they have different spurious correlations.
If a model exploits a shortcut, its performance on the groups that agree with the associated spurious correlation improves, while that on the groups disagreeing with such a spurious correlation deteriorates, leading to \textit{group conflicts}.
We consider a between-group conflict to be an indication that a model is biased towards spurious correlations, which is illustrated in Figure~\ref{fig:idea}.

Then, to mitigate multiple biases during training, our algorithm trains a model by alleviating all between-group conflicts at once so that it performs well across all groups.
This optimization process, derived from a multi-objective optimization (MOO) algorithm of~\citet{desideri2012multiple}, aims for Pareto optimality, \ie, a state where no group can be further improved without sacrificing others.
To be specific, our algorithm optimizes a linear combination of group-wise losses while dynamically adjusting group-wise importance weights so that model parameters converge to
the minimax Pareto solution, which is a Pareto-stationary solution that minimizes the maximum group loss.

We also introduce a new multi-bias benchmark along with the new debiased training algorithm.
Our benchmark, dubbed MultiCelebA, is a collection of real facial images from CelebA~\citep{CelebA}, and incorporates multiple bias types that are spuriously correlated with target classes. 
Compared with existing multi-bias datasets composed of synthetic images~\citep{li2022discover, li2023whac}, it allows to evaluate debiased training algorithm on more realistic and challenging scenarios.

We extensively evaluated our method on three multi-bias benchmarks including MultiCelebA and three single-bias benchmarks, where it outperformed every prior arts.
The main contribution of this paper is four-fold:
\begin{itemize}%
    \setlength\itemsep{-0.5mm}
    \item We present a novel debiased training algorithm based on MOO for mitigating multiple biases simultaneously. 
    \item We present a new real-image multi-bias benchmark for evaluating debiased training methods under realistic and challenging scenarios. 
    \item We benchmarked existing methods for debiased training and demonstrated that they struggle when training data exhibit multiple biases. 
    \item Our method achieved the state of the art on three datasets with multiple biases. In addition, it also showed superior performance on conventional single-bias datasets. 
\end{itemize}

\section{Related Work}

\noindent\textbf{Debiasing in single bias scenarios.}
A body of research has addressed the bias issue that arises from spurious correlations between target and latent attributes.
A group of previous work exploits manual labels for bias attributes~\citep{arjovsky2019invariant,bahng2020learning,dhar2021pass, gong2020jointly, repair,sagawa2019distributionally,teney2020unshuffling,tartaglione2021end,CSAD_2021_ICCV, yao2022improving, zhang2022correct, hex, kirichenko2022last}.
For instance, \citet{sagawa2019distributionally} presented a robust optimization method that weights groups of different bias attributes differently, \citet{dhar2021pass} and \citet{gong2020jointly} employed adversarial training, and \citet{zhang2022correct} proposed using contrastive learning.
Later on, debiased training algorithms that do not require any bias supervision have been studied to reduce the annotation cost~\citep{LAD, creager2021environment, biaswap, LDR, LfF, JTT, ahmed2020systematic, kim2022learning, hwang2022selecmix, zhang2022contrastive, yang2023mitigating, wu2023discover}.
However, whether directly using the bias labels or not, these methods assume that the bias inherent in data is of a single type.
This assumption often does not hold in real-world scenarios, where data exhibit multiple biases, and in practice classifiers can be easily biased to multiple independent biases, as shown in StylEx~\citep{lang2021explaining}.

\textbf{Debiasing in multiple biases scenarios. }
Only a few recent studies~\citep{li2022discover, li2023whac} addressed multiple biases with new training algorithms and benchmarks. 
\citet{li2022discover} discovered multiple biases through iterative assignment of pseudo bias labels, while \citet{li2023whac} presented an augmentation method that emulates the generation process of bias types.
However, their methods are dedicated to handle synthetic images.
In contrast, we propose a new algorithm that trains unbiased models regardless of the number and types of biases, along with a new natural image benchmark for evaluating debiased training methods in the presence of multiple biases.

\section{Proposed Method}
We propose a novel debiased training framework that incorporates a grouping strategy to unveil model biases and an optimization algorithm based on a theory of MOO~\citep{desideri2012multiple}, assuming that bias attributes are annotated for training data. 
This method effectively addresses one or multiple spurious correlations by training a model for all the groups while optimizing importance weights of the groups as well as the model parameter.
The rest of this section first introduces the grouping strategy (Section~\ref{method: grouping}) and then describes the proposed optimization algorithm with group-wise importance weight adjustment in detail (Section~\ref{OurMethod})

\subsection{Grouping Strategy}\label{method: grouping}
As illustrated in Figure~\ref{fig:grouping}(a), we divide training data into multiple groups so that all data in the same group have the same impact on training in terms of the model bias. 
To be specific, we consider training a classifier on a dataset $\mathcal{X}=\{(x^{(m)}, t^{(m)})\}_{m=1}^{M}$,  where each sample $x^{(m)}$ is associated with a target class $t^{(m)}$ and a list of attributes $\boldsymbol{b}^{(m)} = [b_{1}^{(m)},\ldots, b_{D}^{(m)}]^{\top}$, where $D$ is the number of bias types. 
We group the samples using a list of binary group labels $\boldsymbol{g}^{(m)} = [g_{1}^{(m)},\ldots, g_{D}^{(m)}]$ based on whether each attribute $b^{(m)}_{d}$ is the \textit{majority attribute} in target class $t^{(m)}$, \ie, $g_{d}^{(m)} = 1$ if 
\begin{equation*}
b_{d}^{(m)} = \operatorname*{argmax}_{b_{d}} \Big|\big\{m' | t^{(m')} = t^{(m)}, b^{(m')}_{d} = b_{d}\big\}\Big|,
\end{equation*}
and $g_{d}^{(m)} = 0$ otherwise. This results in $2^{D}$ groups where samples in the same group share the same group labels. 

Our grouping policy differs from prior work that uses the target classes and the attributes as the group labels \citep{sagawa2019distributionally,kirichenko2022last,nam2022spread,sagawa2020investigation,zhang2022correct}: each group in our method contains samples from all the target classes, while the existing ones only keep a group of samples with the same target class and the same attributes.
Hence, our grouping policy enables to conduct the target classification task on each group. 
Moreover, since different groups have different combinations of spurious correlations, a model should not rely on any spurious correlation to work on every group;
if it is biased to a spurious correlation, its performance will deteriorate on the groups disagreeing with the spurious correlation.
Our goal in the following debiased training step is thus to train a model capable of accurately classifying samples of all the groups, \ie, its performance should not be biased towards a certain group.

\subsection{Debiased Training with Group Weight Adjustment}\label{OurMethod}

Our debiased training algorithm aims to train a model so that it works for all the groups determined by our grouping policy described in Section~\ref{method: grouping}. To this end, our algorithm optimizes a linear combination of group-wise losses while adjusting their importance weights dynamically. 
In this section, we first briefly review a theory of multi-objective optimization, from which our algorithm stems, and deliver the details of our algorithm.

\subsubsection{Preliminary: Multi-objective optimization}\label{MTLandMOO}

We consider MOO as a problem of optimizing a parameter $\theta$ with respect to a collection of training objectives $L(\theta) = [\mathcal{L}_{1}(\theta),\ldots, \mathcal{L}_{N}(\theta)]^{\top}$. To solve such a problem, MOO frameworks aim at finding a solution that achieves Pareto optimality, \ie, a state where no objective can be improved without sacrificing others.
\begin{definition}[Pareto optimality]
A parameter $\theta^*$ is Pareto-optimal if there exists no other parameter $\theta$ such that $\mathcal{L}_{n}(\theta)\leq\mathcal{L}_{n}(\theta^{\ast})$ for $n=1,\ldots, N$ and $L(\theta)\neq L(\theta^{\ast})$.
\end{definition}

However, finding the Pareto-optimal parameter is intractable for non-convex loss functions like training objectives of deep neural networks. Instead, one may consider using gradient-based optimization to find a parameter satisfying Pareto stationarity \citep{desideri2012multiple}, \ie, a state where a convex combination of objective-wise gradients equals a zero-vector. Pareto stationarity is a necessary condition for Pareto optimality if the objectives in $L(\theta)$ are smooth~\citep{desideri2012multiple}. 
\begin{definition}[Pareto stationarity]
A parameter $\theta^{\ast}$ is Pareto-stationary if there exists an objective-scaling vector $\boldsymbol{\alpha}=[\alpha_{1},\ldots, \alpha_{N}]^{\top}$ satisfying the following condition:
\begin{align}
    \boldsymbol{\alpha}^{\top} \nabla_{\theta} L(\theta^\ast)=\boldsymbol{0}, \quad \boldsymbol{\alpha}\geq \boldsymbol{0}, \quad \boldsymbol{\alpha}^{\top}\boldsymbol{1} = 1, 
\end{align}
where $\boldsymbol{0} = [0,\ldots, 0]^{\top} \in \mathbb{R}^{N}$ and $\boldsymbol{1} = [1,\ldots, 1]^{\top} \in \mathbb{R}^{N}$.
\end{definition} 

\citet{desideri2012multiple} proposed the multi-gradient descent algorithm (MGDA)
to search for a Pareto-stationary parameter.
MGDA finds an objective-scaling parameter $\boldsymbol{\alpha}$ which combines the objective-wise gradients $\nabla_{\theta}L$ to sum to approximately a zero vector by the following optimization:
\begin{align}\label{eqn:MOO}
    \min_{\boldsymbol{\alpha}} \left\lVert \boldsymbol{\alpha}^{\top} \nabla_\theta L\right\rVert_2^2, \quad \boldsymbol{\alpha}\geq \boldsymbol{0}, \quad \boldsymbol{\alpha}^{\top}\boldsymbol{1} = 1.
\end{align}
Given $\boldsymbol{\alpha}$, MGDA performs a gradient-based update on the parameter $\theta$ with respect to $\boldsymbol{\alpha}^{\top}L(\theta)$.

\subsubsection{Proposed Training Algorithm}
We propose an algorithm to optimize over $N=2^{D}$ groups while minimizing the conflicts between group-wise loss functions, inspired by MGDA. 
Let $L(\theta) = [\mathcal{L}_{1}(\theta),\ldots,\mathcal{L}_{N}(\theta)]^{\top}$ denote the list of empirical risk functions on $N$ groups and consider minimizing their convex combination 
$\boldsymbol{\alpha}^{\top}L(\theta)$, where $\boldsymbol{\alpha}\geq \boldsymbol{0}$ and $\boldsymbol{\alpha}^{\top}\boldsymbol{1}=1$.
This is a unique MOO scenario, in which all objectives are of the same loss function but differ in input distribution (\ie, groups).
To impose the constraints on \(\boldsymbol{\alpha}\) in Eq.~(\ref{eqn:MOO}), we propose applying the softmax function $\sigma(\cdot)$ to $\boldsymbol{\alpha}$.
The model parameter $\theta$ is thus optimized by minimizing the weighted group-wise losses:
\begin{figure}
\vspace{-2.5mm}
\begin{algorithm}[H]
\begin{algorithmic}
    \caption{Debiased training by MOO}\label{alg:algorithm}
        \WHILE{not converged}
            \FOR{$u \gets 1$ \text{to} $U-1$}
                \STATE Update $\theta \gets\theta-\eta_1 \nabla_\theta L_{\theta}$.
            \ENDFOR
            
            Update parameters:
            
            \hspace{\algorithmicindent}$\theta \gets\theta-\eta_1 \nabla_\theta L_{\theta}$, 
            
            \hspace{\algorithmicindent}$\boldsymbol{\alpha} \gets \boldsymbol{\alpha} - \eta_2 \nabla_{\boldsymbol{\alpha}} L_{\boldsymbol{\alpha}}$, 
            
            \hspace{\algorithmicindent}$\lambda \gets \lambda + \eta_2 \nabla_{\lambda} L_{\boldsymbol{\alpha}}$.
        \ENDWHILE
\end{algorithmic}
\end{algorithm}
\vspace{-5mm}
\end{figure}

\begin{align}\label{eqn:loss_theta}
    L_{\theta} = \sigma(\boldsymbol{\alpha})^{\top}L(\theta).
\end{align}
To address between-group conflicts, we propose adjusting the group-scaling parameter $\sigma(\boldsymbol{\alpha})$ so that the training converges to a Pareto-stationary solution. 
To be specific, our goal is to minimize the training objective $\sigma(\boldsymbol{\alpha})^{\top}L(\theta)$ while simultaneously adjusting the group-scaling parameter to minimize the objective in Eq.~(\ref{eqn:MOO}). 
To this end, we optimize the following loss function with respect to $\boldsymbol{\alpha}$:
    \begin{align}\label{eqn:loss_alpha}
        L_{\boldsymbol{\alpha}} = \sigma(\boldsymbol{\alpha})^{\top}L(\theta) +\lambda\left\lVert \sigma(\boldsymbol{\alpha})^{\top}(\nabla_\theta L(\theta))_{\dagger}\right\rVert_2^2,
    \end{align}
where $(\cdot)_{\dagger}$ denotes the stop-gradient operator and $\lambda$ is the Lagrangian multiplier for the Pareto stationarity objective in Eq.~(\ref{eqn:MOO}). 
We update the group-scaling parameter with gradient descent and the Lagrangian multiplier $\lambda$ with gradient ascent every $U$ iteration. The learning process of our method is described in Algorithm~\ref{alg:algorithm}. 
Our algorithm encourages a model to achieve the minimax Pareto solution among Pareto-stationary solutions, which minimizes the maximum group loss by emphasizing groups with lower accuracy, \ie, increasing their scaling parameters. This approach enables debiased training across groups. Further details are presented in Section~\ref{sec:discussion_minimax}.

We also note that our method can be interpreted as curvature aware training \citep{li2021robust}, where the group-scaling parameter $\boldsymbol{\alpha}$ is adjusted for better generalization on each group. 
Specifically, \citet{li2021robust} consider adjusting the training objective $\left\lVert \boldsymbol{\alpha}^{\top}(\nabla_\theta L(\theta))\right\rVert_2^2$ so that gradient-based optimization of the parameter $\theta$ converges to a \textit{flat minimum} with a small curvature, \ie, a parameter with a small trace of the Hessian matrix with respect to the training objective. 
It has been reported in the literature that a model converging to such a flat minimum in training has better generalization capability~\citep{keskar2016large,dziugaite2017computing,jiang2019fantastic}.
Since the number of samples that disagree with all the spurious correlations is extremely small and prone to overfitting, 
improving generalization capability is particularly beneficial in multiple biases scenarios.

\subsection{Discussion}\label{sec:discussion}
\subsubsection{Why our algorithm achieves the minimax Pareto solution}\label{sec:discussion_minimax}
During training, our algorithm more emphasizes groups with worse accuracy by increasing their scaling parameters, which encourages achieving the minimax Pareto solution. This behavior of our algorithm is attributed to the following two factors of $L_{\boldsymbol{\alpha}}$ in Eq.~(\ref{eqn:loss_alpha}).

\textbf{Regarding the first term of $L_{\boldsymbol{\alpha}}$}: Minimizing this term substantially increases the group-scaling parameter of the CC group that exhibits the worst performance in testing, consequently improving the worst accuracy. 
To be specific, as $\sigma(\boldsymbol{\alpha})$ holds the sum-to-one constraint, minimizing the first term increases the scaling parameter for groups with lower loss magnitudes while decreasing the parameter for groups with higher loss magnitudes. 
Here, the CC group shows the lowest training loss scale since its cardinality is extremely small and the model is easily overfitted to the group (which causes the worst accuracy on the group in testing). Further empirical analysis on this term is provided in Appendix~\ref{supp_section:loss}.

\textbf{Regarding the second term of $L_{\boldsymbol{\alpha}}$}: This term, originated from MGDA, also increases the scaling parameter for the CC group. MGDA is known to be biased towards tasks with low loss magnitudes in multi-task learning, a phenomenon known as \textit{task impartiality}~\citep{javaloy2021rotograd, liu2021towards}. In our setting, the task with the lowest loss magnitude corresponds to the CC group as discussed above. Hence, the task impartiality of MGDA leads to the increased scaling parameter for the CC group, leading to the minimax Pareto solution.

\subsubsection{Comparison with Multi-task Learning}
Multi-task learning (MTL) is a research area that aims to develop a model capable of performing multiple tasks simultaneously. Considering that MTL incorporates MOO to address task conflicts~\citep{sener2018multi}, 
MTL exhibits similarities with our method.
Nevertheless, our work is clearly distinct from MTL in multiple aspects. 
Firstly, our work addresses a single classification task and thus the group-wise losses have the same form. However, their input distributions differ, with each group-wise loss calculated using samples from its respective group.
In contrast, MTL assumes different loss functions for different tasks.
Secondly, since all the groups aim to solve the same target task, an optimal solution that fits perfectly across all the groups exists for our debiased training setting in principle. 
On the contrary, MTL rarely has the perfect solution for all the tasks since task conflicts are almost inevitable.
Third, our method does not employ task-specific network parameters unlike MTL, which in general distinguishes task-specific and task-shared parameters.
Lastly, we present a novel loss tailored to the debiased training.

\subsubsection{On the Use of Bias Labels}
Bias attribute labels would be expensive, particularly in the multi-bias setting. 
However, regarding that debiasing in this setting has been rarely studied so far and is extremely challenging, we believe it is premature to tackle the task in an unsupervised fashion at this time.
As in the single-bias setting where the society has first developed supervised debiasing methods and then unsupervised counterparts, our algorithm will be a cornerstone of follow-up unsupervised methods in the multi-bias setting. 
Moreover, the annotation cost for bias labels can be substantially reduced by incorporating existing techniques for pseudo labeling of bias attributes~\citep{jung2021learning,nam2022spread}.

\begin{figure}[t]
\centering
\includegraphics[width=\linewidth]{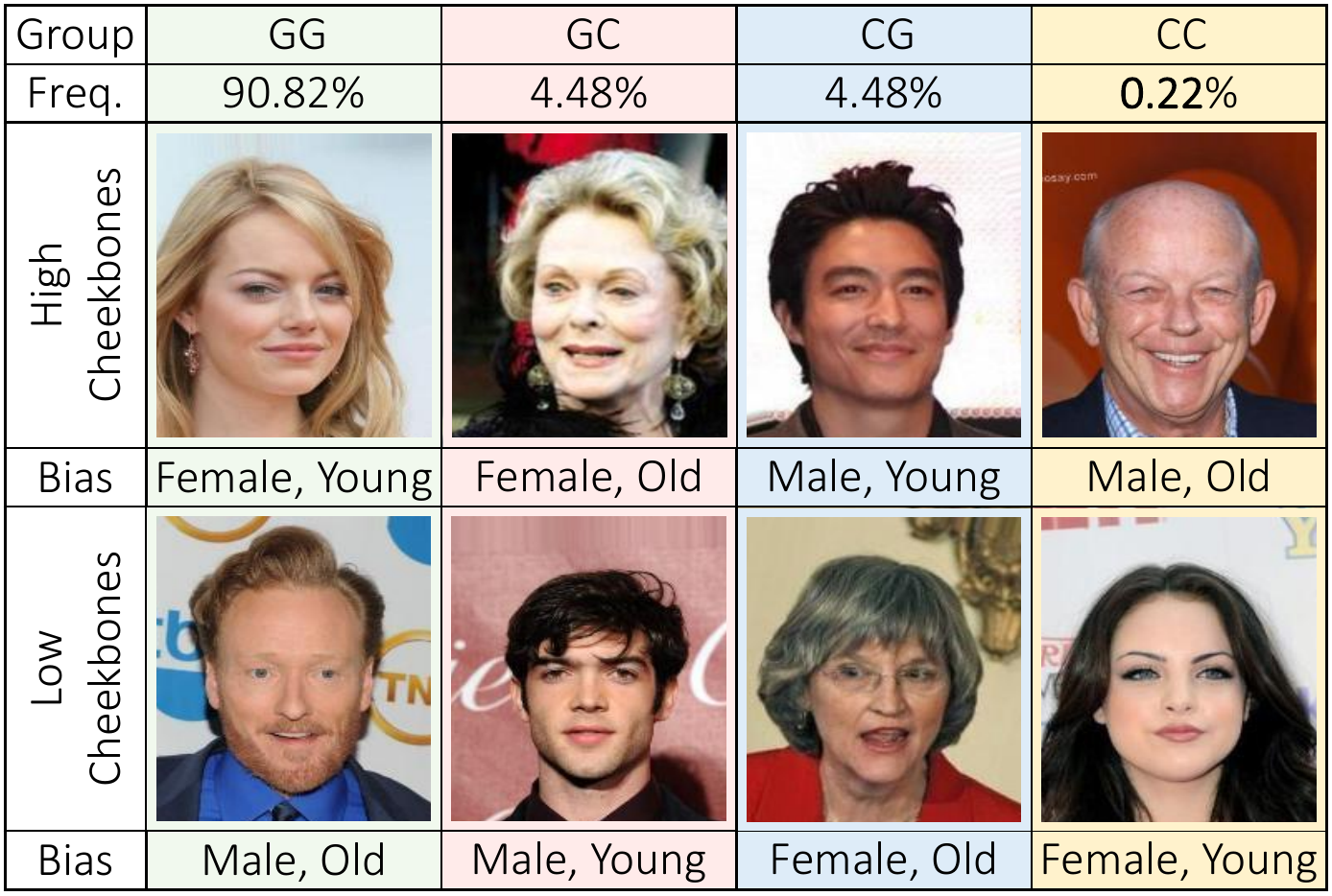}

\caption{
Training set configuration of MultiCelebA in the two-bias setting.
}
\label{fig:MultiCeleba_examples}
\end{figure}

\section{MultiCelebA Benchmark} \label{MultiCelebA}
We present a new benchmark, dubbed MultiCelebA, for evaluating debiased training algorithms under the presence of multiple biases.  
Unlike Multi-Color MNIST~\citep{li2022discover} and UrbanCars~\citep{li2023whac} built for the same purpose using synthetic images, MultiCelebA is composed of natural facial images, making it more suitable for simulating real-world scenarios.

MultiCelebA is built upon CelebA~\citep{CelebA}, a large-scale collection of facial images each with 40 attribute annotations. 
Among these attributes, \texttt{high-cheekbones} is chosen as the target class, while \texttt{gender}, \texttt{age}, and \texttt{mouth slightly open} are used as bias attributes that are spuriously correlated with \texttt{high-cheekbones} and thus cause undesirable shortcuts during training.
Note that these bias attributes are not randomly chosen but identified by adapting the empirical analysis procedure of \citet{scimeca2021shortcut} to CelebA, which revealed that these attributes are strongly correlated with the target class; details of the analysis are presented in Appendix~\ref{supp_section:diagonal}.

Based on MultiCelebA, we present two different benchmark settings: one with two bias attributes \texttt{gender} and \texttt{age}, and the other with all three bias attributes. 
To simulate challenging scenarios where training data are extremely biased, we set the bias-guiding samples for both \texttt{gender} and \texttt{age} to $95.3\%$ by subsampling from the CelebA training set, so that only 0.22\% of training samples are free from spurious correlations in the two-bias setting and 0.07\% for the three-bias settings.
Example images and the frequency of each attribute in the two-bias setting are presented in Figure~\ref{fig:MultiCeleba_examples}.

\section{Experiments} \label{Experiments}

\begin{table*}[t]
\caption{
Performance in \textsc{InDist}, \textsc{GG}, \textsc{GC}, \textsc{CG}, \textsc{CC}, and \textsc{Unbiased} (\%) on MultiCelebA in two-bias setting.
The first element of each of the four combinations \{\textsc{GG}, \textsc{GC}, \textsc{CG}, \textsc{CC}\} is about the bias type \texttt{gender}, while the second is about the bias type \texttt{age}.
We mark the best and the second-best performance in \textbf{bold} and \underline{underline}, respectively.
}
\vspace{-2mm}
\begin{center}
\begin{small}
\begin{tabular}{lc|cccccc}
\hline
Method & Bias label & \textsc{InDist} &\textsc{GG} &\textsc{GC} &\textsc{CG} &\textsc{CC} & \textsc{Unbiased}\\ %
\hline
ERM & \bxmark  & \textbf{97.0}$_{\pm \text{{0.2}}}$ & \textbf{98.2}$_{\pm \text{{0.7}}}$ & \textbf{89.2}$_{\pm \text{{2.6}}}$ & 58.2$_{\pm \text{3.0}}$ & 19.0$_{\pm \text{1.8}}$ & 63.8$_{\pm \text{1.2}}$\\
LfF &\bxmark  & 81.9$_{\pm \text{3.1}}$ & 79.8$_{\pm \text{2.6}}$ & 71.7$_{\pm \text{2.2}}$ & 80.2$_{\pm \text{1.7}}$ & 71.5$_{\pm \text{3.3}}$ & 75.8$_{\pm \text{0.5}}$\\
JTT &\bxmark &78.7$_{\pm \text{6.5}}$ & 76.1$_{\pm \text{5.2}}$ &60.8$_{\pm \text{5.2}}$ &65.1$_{\pm \text{10.7}}$ &51.9$_{\pm \text{1.6}}$ &64.7$_{\pm \text{3.2}}$\\
DebiAN & \bxmark & 66.8$_{\pm \text{34.1}}$ & 64.4$_{\pm \text{30.4}}$ & 63.6$_{\pm \text{22.2}}$ & 49.8$_{\pm \text{7.6}}$ & 45.5$_{\pm \text{13.2}}$ & 55.8$_{\pm \text{11.7}}$ \\
Upsampling & \bcmark & 82.6$_{\pm \text{0.8}}$ & 79.8$_{\pm \text{1.5}}$ &81.0$_{\pm \text{1.3}}$ & 76.7$_{\pm \text{1.1}}$ & 75.6$_{\pm \text{1.2}}$ & 78.3$_{\pm \text{0.8}}$ \\
Upweighting & \bcmark & 83.4$_{\pm \text{5.9}}$ & 79.0$_{\pm \text{4.1}}$ & 79.2$_{\pm \text{6.0}}$ & \underline{80.8}$_{\pm \text{0.0}}$ & \underline{78.7}$_{\pm \text{3.6}}$ & 79.4$_{\pm \text{3.4}}$\\
GroupDRO &\bcmark & 83.5$_{\pm \text{0.7}}$ & 81.2$_{\pm \text{1.0}}$ & 81.2$_{\pm \text{1.2}}$ & 76.7$_{\pm \text{1.5}}$ & 74.6$_{\pm \text{0.4}}$ & 78.4$_{\pm \text{0.7}}$ \\
SUBG& \bcmark & 80.3$_{\pm \text{1.1}}$ & 77.1$_{\pm \text{1.0}}$ & 78.4$_{\pm \text{0.7}}$ & 77.5$_{\pm \text{1.7}}$ & 78.0$_{\pm \text{1.2}}$ & 77.7$_{\pm \text{0.6}}$\\
LISA & \bcmark & 84.5$_{\pm \text{1.7}}$ & 82.8$_{\pm \text{1.3}}$ & 83.2$_{\pm \text{0.5}}$ & 79.8$_{\pm \text{0.8}}$ & 77.6$_{\pm \text{2.6}}$ & \underline{80.9}$_{\pm \text{0.2}}$\\
DFR$_{tr}^{tr}$ & \bcmark & \underline{85.5}$_{\pm \text{6.2}}$ & \underline{91.3}$_{\pm \text{3.5}}$ & 83.6$_{\pm \text{4.0}}$ & 46.7$_{\pm \text{3.8}}$ & 28.5$_{\pm \text{4.6}}$ & 62.5$_{\pm \text{0.6}}$\\ 
\rowcolor{Gray}
Ours &\bcmark & 84.3$_{\pm \text{0.9}}$ & 82.4$_{\pm \text{0.9}}$ & \underline{85.1}$_{\pm \text{0.4}}$ & \textbf{81.7}$_{\pm \text{0.4}}$ & \textbf{82.6}$_{\pm \text{1.0}}$ & \textbf{82.9}$_{\pm \text{0.2}}$ \\ %
\hline
\end{tabular}
\end{small}
\end{center}
\label{tab:multiceleba_table}
\end{table*}

\begin{table*}[t]
    \begin{minipage}{0.49\textwidth}
        \caption{
        Performance in \textsc{InDist}, \textsc{CCC}, and \textsc{Unbiased} (\%) on MultiCelebA in three biases for evaluation two biases for training setting. We mark the best and the second-best performance in \textbf{bold} and \underline{underline}, respectively.
        }
        \vspace{-2mm}
        \centering
        \begin{small}
        
        \begin{tabular}{l|ccc}
        \hline
        Method  & \textsc{InDist} & \textsc{CCC} &\textsc{Unbiased} \\ %
        \hline
        ERM  & \textbf{96.7}$_{\pm \text{0.2}}$ & 11.1$_{\pm \text{2.8}}$ & $63.8_{\pm \text{1.2}}$ \\
        LfF & 81.8$_{\pm \text{3.1}}$ & 60.1$_{\pm \text{1.7}}$ & 71.8$_{\pm \text{0.7}}$ \\ 
        Upsampling & 82.6$_{\pm \text{0.8}}$ & 63.0$_{\pm \text{3.5}}$ & 73.6$_{\pm \text{0.8}}$\\ 
        Upweighting & 85.4$_{\pm \text{9.5}}$ & 63.4$_{\pm \text{3.5}}$ & 75.8$_{\pm \text{4.8}}$ \\
        GroupDRO & 83.4$_{\pm \text{0.6}}$ & 61.4$_{\pm \text{2.9}}$ & 73.7$_{\pm \text{0.7}}$\\
        SUBG & 80.3$_{\pm \text{1.1}}$ & \underline{65.8}$_{\pm \text{4.0}}$ & 72.6$_{\pm \text{1.3}}$ \\
        LISA & 84.5$_{\pm \text{1.7}}$ & 63.1$_{\pm \text{1.0}}$ & \underline{75.9}$_{\pm \text{0.6}}$  \\
        DFR$_{tr}^{tr}$ & \underline{85.5}$_{\pm \text{6.1}}$ & 26.2$_{\pm \text{6.2}}$ & 61.3$_{\pm \text{0.6}}$\\
        \rowcolor{Gray}
        Ours & 84.3$_{\pm \text{1.0}}$ & \textbf{70.0}$_{\pm \text{0.6}}$ & \textbf{78.4}$_{\pm \text{0.0}}$ \\
        \hline
        \end{tabular}
        \end{small}
        \label{tab:3types_table}
    \end{minipage}
    \hfill
    \begin{minipage}{0.49\textwidth}
        \caption{Performance in \textsc{InDist} and \textsc{CC} (\%) on UrbanCars.
        We mark the best and the second-best in \textbf{bold} and \underline{underline}, respectively. 
        }
        \vspace{-2mm}
        \centering
        \begin{small}
        \begin{tabular}{lc|ccc}
        \hline
        Method & Bias label & \textsc{InDist} & \textsc{CC} & \textsc{Gap}\\
        \hline
        ERM         & \bxmark   & \textbf{97.6} & 28.4 & -69.2\\
        Upsampling  & \bcmark   & 92.8  & 76.0 & -16.8 \\
        Upweighting & \bcmark   & 93.4  & 80.0 & -13.4 \\ 
        GroupDRO    & \bcmark   & 91.6  & 75.2 & -16.4\\
        SUBG        & \bcmark   & 71.1  & 64.8 & \underline{-6.3}\\
        LISA        & \bcmark   & \underline{94.6}  & \underline{80.8} & -13.8\\
        DFR$_{tr}^{tr}$& \bcmark   & 89.7  & 44.5 & -45.2\\
        \rowcolor{Gray}
        Ours        & \bcmark   & 91.8 & \textbf{87.6} & \textbf{-4.2} \\
        \hline
        \end{tabular}
        \end{small}
        \label{tab:urbancars}
    \end{minipage}
\end{table*}

\subsection{Setup}

\noindent\textbf{Datasets.} We adopt three multi-bias benchmarks, MultiCelebA, UrbanCars~\citep{li2023whac}, and Multi-Color MNIST~\citep{li2022discover}, and three single-bias datasets, Waterbirds~\citep{sagawa2019distributionally}, CelebA~\citep{CelebA}, and BFFHQ~\citep{LDR} for evaluation.

\noindent\textbf{Evaluation metrics.} 
For multi-bias benchmarks, the quality of debiased training algorithms is measured mainly by \textsc{Unbiased}, the average of group average accuracy scores. For the benchmarks with two bias types, we also adopt average accuracy for each of the four groups categorized by the guiding or conflicting nature of the biases: \{\textsc{GG}, \textsc{GC}, \textsc{CG}, \textsc{CC}\}, where \textsc{G} and \textsc{C} indicate whether a group includes bias-guiding or bias-conflicting samples for each bias type, respectively. 
Conceptually, the \textsc{GG} metric can be high regardless of whether a model is biased or not. 
However, the \textsc{CC} metric can be high only when a model is debiased from all spurious correlations.
Similarly, the \textsc{GC} accuracy can be high only when a model is debiased from the spurious correlation of the second bias type.
We also report \textsc{InDist}, the weighted average of group accuracy scores where the weights are proportional to group sizes of training data \citep{sagawa2019distributionally}.
For Waterbirds and CelebA, we adopt \textsc{Worst}, the minimum of group accuracy scores, following \citet{sagawa2019distributionally}.

\begin{table*}[t!]

\caption{Performance in \textsc{GG}, \textsc{GC}, \textsc{CG}, \textsc{CC}, and \textsc{Unbiased} ($\%$) on Multi-Color MNIST.
The first element of each of the four combinations \{\textsc{GG}, \textsc{GC}, \textsc{CG}, \textsc{CC}\} is about the bias type \texttt{left-color}, while the second is about the bias type \texttt{right-color}.
We mark the best and the second-best performance in \textbf{bold} and \underline{underline}, respectively.
}
\vspace{-2mm}
\begin{center}
\begin{small}
\begin{tabular}{lc|ccccc}
\hline
Method & Bias label &\textsc{GG} &\textsc{GC} &\textsc{CG} &\textsc{CC} & \textsc{Unbiased}\\ %
\hline
ERM &\bxmark & 100.0$_{\pm \text{0.0}}$	&\underline{96.5}$_{\pm \text{1.2}}$ &79.5$_{\pm \text{2.5}}$	&20.8$_{\pm \text{1.1}}$ & 74.2$_{\pm \text{1.1}}$ \\ %
LfF &\bxmark &99.6$_{\pm \text{0.5}}$ &4.7$_{\pm \text{0.5}}$ &\textbf{98.6}$_{\pm \text{0.4}}$ &5.1$_{\pm \text{0.4}}$ &52.0$_{\pm \text{0.1}}$ \\
EIIL &\bxmark &100.0$_{\pm \text{0.0}}$ & \textbf{97.2}$_{\pm \text{1.5}}$ &70.8$_{\pm \text{4.9}}$ &10.9$_{\pm \text{0.8}}$ &69.7$_{\pm \text{1.0}}$\\
PGI & \bxmark &98.6$_{\pm \text{2.3}}$ &82.6$_{\pm \text{19.6}}$ &26.6$_{\pm \text{5.5}}$ &9.5$_{\pm \text{3.2}}$ &54.3$_{\pm \text{4.0}}$ \\
DebiAN & \bxmark &100.0$_{\pm \text{0.0}}$ &95.6$_{\pm \text{0.8}}$ &76.5$_{\pm \text{0.7}}$ &16.0$_{\pm \text{1.8}}$ &72.0$_{\pm \text{0.8}}$\\
Upsampling & \bcmark & 99.4$_{\pm \text{0.6}}$ &89.8$_{\pm \text{1.4}}$  &81.3$_{\pm \text{2.6}}$  &42.0$_{\pm \text{1.7}}$  &78.1$_{\pm \text{1.4}}$ \\
Upweighting& \bcmark & 100.0$_{\pm \text{0.0}}$ &90.0$_{\pm \text{2.5}}$  & \underline{83.4}$_{\pm \text{2.1}}$  &37.1$_{\pm \text{2.8}}$  &77.6$_{\pm \text{1.0}}$ \\
GroupDRO &\bcmark & 98.0$_{\pm \text{0.0}}$ &87.2$_{\pm \text{4.3}}$ &77.3$_{\pm \text{7.5}}$ & \textbf{52.3}$_{\pm \text{2.6}}$ &\underline{78.7}$_{\pm \text{2.7}}$ \\
\rowcolor{Gray}
Ours &\bcmark & 99.7$_{\pm \text{0.6}}$	&90.4$_{\pm \text{3.4}}$	& 81.8$_{\pm \text{4.0}}$ & \underline{48.1}$_{\pm \text{0.3}}$ & \textbf{80.0}$_{\pm \text{2.0}}$ %
\\
\hline
\end{tabular}
\end{small}
\end{center}
\label{tab:multi-color-mnist}
\end{table*}

\begin{table*}[h]
\begin{minipage}[t]{0.65\textwidth}
\caption{
\textsc{Worst} and \textsc{InDist} metrics (\%) evaluated on Waterbirds, and CelebA. We mark the best and the second-best performance of \textsc{Worst} in \textbf{bold} and \underline{underline}, respectively. 
}
\vspace{-2mm}
\begin{center}
\begin{small}
\begin{tabular}{lc|cccc}
\hline
&Bias&\multicolumn{2}{c}{Waterbirds} & \multicolumn{2}{c}{CelebA}\\
Method & label & \textsc{Worst} & \textsc{InDist} & \textsc{Worst} & \textsc{InDist}\\
\hline
ERM & \bxmark & 63.7$_{\pm \text{1.9}}$ & 97.0$_{\pm \text{0.2}}$ & 47.8$_{\pm\text{3.7}}$ & 94.9$_{\pm \text{0.2}}$\\
LfF &\bxmark & 78.0\phantom{$_{\pm \text{0.0}}$} & 91.2\phantom{$_{\pm \text{0.0}}$} & 70.6\phantom{$_{\pm \text{0.0}}$} & 86.0\phantom{$_{\pm \text{0.0}}$}\\ %
EIIL &\bxmark & 77.2$_{\pm \text{1.0}}$ & 96.5$_{\pm \text{0.2}}$ & 81.7$_{\pm \text{0.8}}$ & 85.7$_{\pm \text{0.1}}$\\
JTT &\bxmark  & 83.8$_{\pm \text{1.2}}$ & 89.3$_{\pm \text{0.7}}$ & 81.5$_{\pm \text{1.7}}$ & 88.1$_{\pm \text{0.3}}$\\
LWBC & \bxmark & - & - & 85.5$_{\pm \text{1.4}}$ & 88.9$_{\pm \text{1.6}}$\\
CNC & \bxmark & 88.5$_{\pm \text{0.3}}$ & 90.9$_{\pm \text{0.1}}$  & 88.8$_{\pm \text{0.9}}$ & 89.9$_{\pm \text{0.5}}$\\ 
Upweighting & \bcmark & 88.0$_{\pm \text{1.3}}$ & 95.1$_{\pm \text{0.3}}$ & 83.3$_{\pm \text{2.8}}$ & 92.9$_{\pm \text{0.2}}$\\ %
GroupDRO &\bcmark & 89.9$_{\pm \text{0.6}}$ & 92.0$_{\pm \text{0.6}}$ & 88.9$_{\pm \text{1.3}}$ & 93.9$_{\pm \text{0.1}}$\\ %
SUBG & \bcmark & 89.1$_{\pm \text{1.1}}$ & - & 85.6$_{\pm \text{2.3}}$ & - \\
SSA & \bcmark & 89.0$_{\pm \text{0.6}}$ & 92.2$_{\pm \text{0.9}}$ & \textbf{89.8}$_{\pm \text{1.3}}$ & 92.8$_{\pm \text{0.1}}$ \\
LISA & \bcmark  & 89.2$_{\pm \text{0.6}}$ & 91.8$_{\pm \text{0.3}}$ & \underline{89.3}$_{\pm \text{1.1}}$ & 92.4$_{\pm \text{0.4}}$\\
DFR$_{tr}^{tr}$ & \bcmark  & \underline{90.2}$_{\pm \text{0.8}}$ & 97.0$_{\pm \text{0.3}}$ & 80.7$_{\pm \text{2.4}}$ & 90.6$_{\pm \text{0.7}}$\\ 
\rowcolor{Gray}
Ours &\bcmark & \textbf{91.8}$_{\pm \text{0.3}}$ & 95.6$_{\pm \text{0.3}}$ & \textbf{89.8}$_{\pm \text{1.3}}$ & 91.4$_{\pm \text{1.2}}$\\
\hline
\end{tabular}
\end{small}
\end{center}
\label{tab:single_bias_setting_main}
\end{minipage}
\hfill
\begin{minipage}[t]{0.32\textwidth}
\caption{
\textsc{Unbiased} metric (\%) evaluated on BFFHQ. We mark the best and the second-best performance in \textbf{bold} and \underline{underline}, respectively. 
}
\vspace{-2mm}
\begin{center}
\begin{small}
\begin{tabular}{l|ccc}
\hline
&Bias& BFFHQ\\
Method & label & \textsc{Unbiased} \\
\hline
ERM &\bxmark & 56.2$_{\pm \text{0.4}}$\\
HEX & \bxmark& 52.8$_{\pm \text{0.9}}$\\
ReBias &\bxmark & 56.8$_{\pm \text{1.6}}$\\
LfF &\bxmark & 65.6$_{\pm \text{1.4}}$\\
DisEnt &\bxmark & 61.6$_{\pm \text{2.0}}$\\
SelecMix & \bxmark &71.6$_{\pm \text{1.9}}$\\ 
SelecMix & \bcmark &75.0$_{\pm \text{0.5}}$\\
EnD &\bcmark & 56.5$_{\pm \text{0.6}}$\\
LISA &\bcmark & 65.2$_{\pm \text{0.5}}$\\
GroupDRO & \bcmark & \underline{85.1}$_{\pm \text{0.9}}$\\
\rowcolor{Gray}
Ours &\bcmark & \textbf{85.7}$_{\pm \text{0.3}}$\\
\hline
\end{tabular}
\end{small}
\end{center}
\label{tab:bffhq_main}
\end{minipage}
\vspace{-2mm}
\end{table*}

\noindent\textbf{Baselines.} 
We compare our algorithm with a large body of existing debiased training algorithms.
Among them, GroupDRO~\citep{sagawa2019distributionally}, EnD~\citep{tartaglione2021end}, SUBG~\citep{sagawa2020investigation}, LISA~\citep{yao2022improving}, and DFR~\cite{kirichenko2022last} as well as simple upsampling and upweighting strategies demand true bias labels of training data like ours,
while HEX~\citep{wang2018learning}, ReBias~\citep{bahng2020learning}, LfF~\citep{LfF}, JTT~\citep{JTT}, EIIL~\citep{creager2021environment}, PGI~\citep{ahmed2020systematic},  DisEnt~\citep{LDR}, LWBC~\citep{kim2022learning}, SelecMix~\citep{hwang2022selecmix},  CNC~\citep{zhang2022correct}, and DebiAN~\citep{li2022discover} do not.

\noindent\textbf{Implementation details.}
Following previous work, we conduct experiments using different neural network architectures for different datasets: a three-layered MLP for Multi-Color MNIST and ResNet18 for MultiCelebA and BFFHQ, ResNet50 for UrbanCars, Waterbirds, and CelebA.
The group-scaling parameter $\boldsymbol{\alpha}$ is initialized to $\frac1N\boldsymbol{1}$ where $N$ is the number of groups, and the Lagrangian multiplier $\lambda$ is initialized to 0.
For mini-batch construction during training, group-balanced sampling is used to compute each loss for multiple groups.
For MultiCelebA, we tuned hyperparameters in the two-bias setting and performed both training and evaluation in the two-bias setting, and conducted evaluation only in the three-bias setting without training.
We report the average and standard deviation of each metric calculated from three runs with different random seeds.
More implementation details are provided in Appendix~\ref{supp_section:details}.

\subsection{Quantitative Results} 

\noindent\textbf{MultiCelebA in two-bias setting.} In Table~\ref{tab:multiceleba_table}, we present the results of our experiments evaluating the performance of various baselines and existing debiased training methods on MultiCelebA.
One can observe how our method outperforms the baselines by a significant margin in \textsc{Unbiased}, \textsc{CG}, and \textsc{CC} metrics. Our method even achieves a second-best accuracy in the \textsc{GC} metric and a moderate accuracy in the \textsc{GG} metric. This highlights how our method successfully prevents performance degradation by simultaneously removing multiple spurious correlations. 
Intriguingly, we observe that algorithms like JTT, DebiAN, and DFR exhibit \textsc{Unbiased} metric similar to or even lower than the vanilla \textsc{ERM} algorithm. Our hypothesis is that this performance degradation stems from conflicts between the removal of different spurious correlations. %
To be specific, JTT~\citep{JTT} exhibits varying accuracy across the \textsc{GG}, \textsc{GC}, \textsc{CG}, and \textsc{CC} groups, indicating that the model is biased towards both $\texttt{gender}$ and $\texttt{age}$ biases. DebiAN~\citep{li2022discover} shows high accuracy in the \textsc{GG} and \textsc{GC} groups, but low accuracy in the \textsc{CG} and \textsc{CC} groups, indicating that the algorithm partially mitigates $\texttt{age}$ bias but still suffers from $\texttt{gender}$ bias. 
We also observe that DFR~\citep{kirichenko2022last} achieves lower \textsc{CC} and \textsc{CG} metrics than ERM, suggesting that an ERM-based feature representation alone is insufficient in multi-bias setting.
The remaining algorithms, \eg, Upsampling, GroupDRO~\citep{sagawa2019distributionally}, and LISA~\citep{yao2022improving} show overall decent performance, but the \textsc{GG} and \textsc{GC} metrics are slightly higher than that in \textsc{CG} and \textsc{CC} groups, indicating that the model is still biased towards the $\texttt{gender}$ attribute.
Surprisingly, the upweighting baseline achieved the second-best performance in \textsc{CG} and \textsc{CC} metrics on MultiCelebA, surpassing all the existing debiased training methods.

\noindent\textbf{MultiCelebA in three-bias setting.}
Results of the evaluation with three bias types are reported in Table~\ref{tab:3types_table}, where only \texttt{gender} and \texttt{age} labels are visible during training. ERM exhibits lower \textsc{CCC} accuracy in the three-bias setting compared to the two-bias setting. This arises as the number of bias types increases, resulting in a substantially reduced size of the smallest group, demonstrating a more challenging setting. In contrast, our method substantially outperformed existing methods and baselines in \textsc{Unbiased} and \textsc{CCC}. This demonstrates the scalability of our method to more than two bias types.

\begin{table*}[t!]
\caption{Comparisons among different strategies for adjusting the group-scaling parameter $\boldsymbol{\alpha}$ on MultiCelebA in two biases setting.
(a) Fixing $\sigma(\boldsymbol{\alpha})$ by $\frac1N\boldsymbol{1}$. (b) Minimizing $\sigma(\boldsymbol{\alpha})^{\top}L(\theta)$.  (c) MGDA. (d) GradNorm. (e) MoCo, the latest MOO method. (f) Ours minimizing $L_{\boldsymbol{\alpha}}$.
}
\begin{center}
\begin{small}
\begin{tabular}{l|cccccc}
\hline
& \textsc{InDist} & \textsc{GG} & \textsc{GC} & \textsc{CG} & \textsc{CC} & \textsc{Unbiased}\\ %
\hline
(a) No optimization & 78.5$_{\pm \text{5.7}}$ & 79.6$_{\pm \text{2.9}}$ & 80.0$_{\pm \text{2.2}}$ & 79.0$_{\pm \text{1.9}}$ & 78.4$_{\pm \text{1.3}}$ & 79.2$_{\pm \text{1.4}}$ \\ %
(b) Minimizing group losses & 81.3$_{\pm \text{3.6}}$ & 76.4$_{\pm \text{2.2}}$ & 77.8$_{\pm \text{0.4}}$ & 77.1$_{\pm \text{2.2}}$ & 78.0$_{\pm \text{1.7}}$ & 77.3$_{\pm \text{0.6}}$ \\ 
(c) MGDA & 82.7$_{\pm \text{3.7}}$ & 81.6$_{\pm \text{3.5}}$& \underline{85.1}$_{\pm \text{2.1}}$ & \underline{80.1}$_{\pm \text{1.3}}$ & \underline{82.3}$_{\pm \text{3.2}}$ & \underline{82.3}$_{\pm \text{0.4}}$ \\
(d) GradNorm & \textbf{86.5}$_{\pm \text{5.4}}$ & \textbf{85.9}$_{\pm \text{5.8}}$ & \textbf{86.9}$_{\pm \text{2.4}}$ & 78.1$_{\pm \text{3.5}}$ & 76.6$_{\pm \text{6.5}}$ & 81.9$_{\pm \text{0.6}}$ \\
(e) MoCo & 83.8$_{\pm \text{1.6}}$ & 81.7$_{\pm \text{1.3}}$ & 81.8$_{\pm \text{2.6}}$ & 77.2$_{\pm \text{0.9}}$ &74.9$_{\pm \text{1.2}}$ & 78.9$_{\pm \text{1.5}}$ \\
\rowcolor{Gray}
(f) Ours & \underline{84.3}$_{\pm \text{0.9}}$ & \underline{82.4}$_{\pm \text{0.9}}$ & \underline{85.1}$_{\pm \text{0.4}}$ & \textbf{81.7}$_{\pm \text{0.4}}$ & \textbf{82.6}$_{\pm \text{1.0}}$ & \textbf{82.9}$_{\pm \text{0.2}}$ \\ 
\hline
\end{tabular}
\label{loss_ablation_table}
\end{small}
\end{center}
\end{table*}

\begin{figure*}[t]
\begin{minipage}{0.49\linewidth}
\centering
\includegraphics[width=\linewidth]{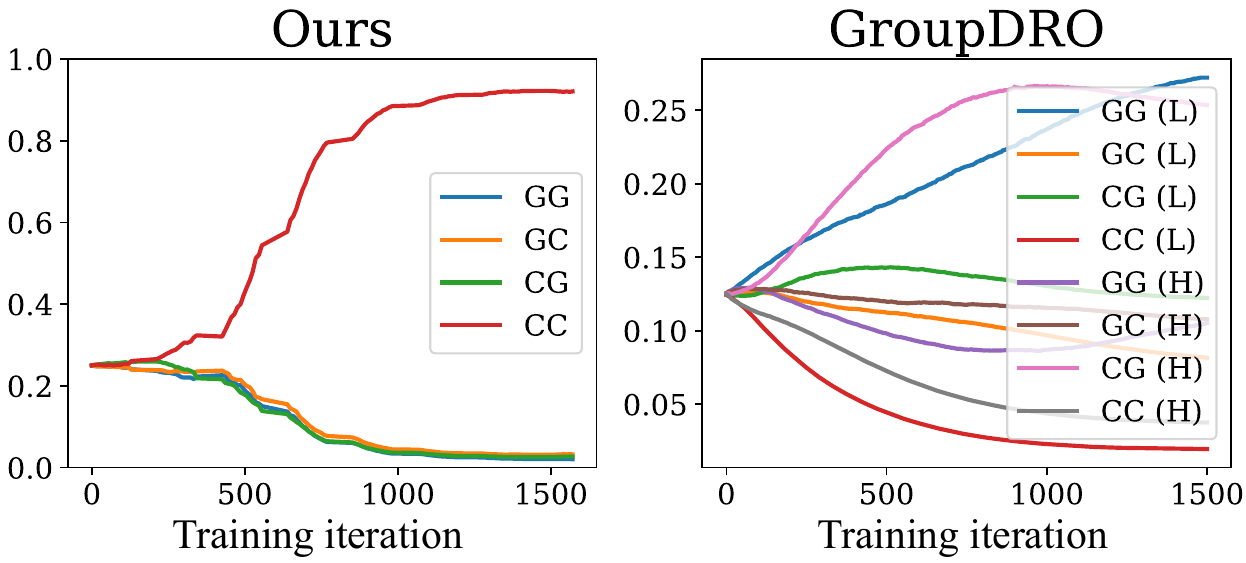}
\caption{Change of the group-scaling parameter over time
on MultiCelebA in two-bias settings.
In the case of GroupDRO, (H) and (L) denote \texttt{High-cheekbones} and \texttt{Low-cheekbones}, respectively.
}
\label{fig:alpha_graph}
\end{minipage}
\hfill
\begin{minipage}{0.49\linewidth}
\centering
\includegraphics[width=\linewidth]{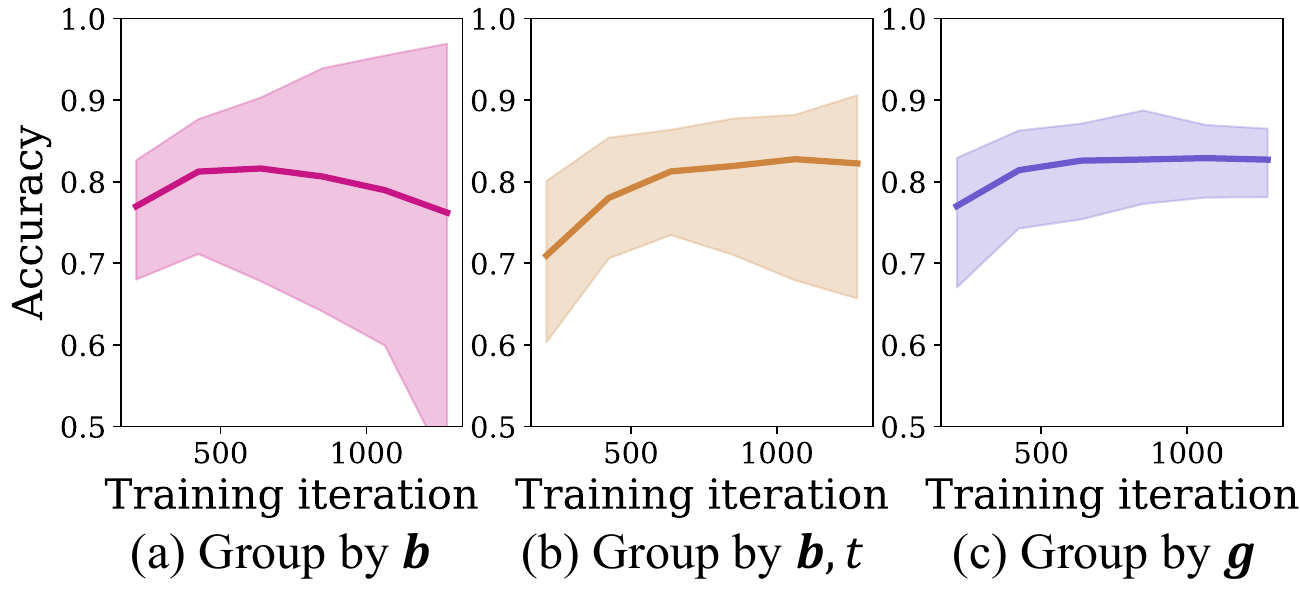}
\caption{
Group-wise test accuracy of three different grouping strategies.
Lines indicate \textsc{Unbiased} performance, and shaded regions show the lowest and the highest accuracy among the group-wise scores. 
To ensure fair comparison, the test data are grouped by $\textbf{b}$ and $t$.
}
\label{fig:ablation_grouping}
\end{minipage}
\end{figure*}

\noindent\textbf{UrbanCars.} In Table~\ref{tab:urbancars}, we present the results of debiased training algorithms that exploit bias labels and share the identical network structure. Our method achieved significantly superior \textsc{CC} accuracy when compared to methods using bias labels, demonstrating a substantial difference.

\noindent\textbf{Multi-Color MNIST.}
In Table~\ref{tab:multi-color-mnist}, we report the evaluation results for the Multi-Color MNIST dataset. Note that we re-use the performance of LfF~\citep{LfF}, EIIL~\citep{creager2021environment}, PGI~\citep{ahmed2020systematic}, and DebiAN~\citep{li2022discover} reported by \citet{li2022discover}. Overall, our method demonstrates the best performance along with GroupDRO. In particular, our algorithm exhibits the highest \textsc{Unbiased} accuracy and the second-best \textsc{CC} accuracy.

\noindent\textbf{Single-bias datasets.} 
In Table~\ref{tab:single_bias_setting_main} and \ref{tab:bffhq_main}, our method achieves the best \textsc{Worst} accuracy on Waterbirds and CelebA, and the best \textsc{Unbiased} on BFFHQ, indicating that our method is effective not only for multi-bias settings but also for single-bias settings.

\subsection{In-depth Analysis}

\noindent\textbf{Comparisons among different strategies for adjusting the group-scaling parameter.} 
We first verify the impact of our strategy for adjusting 
the group-scaling parameter. 
In Table~\ref{loss_ablation_table}, we compare our training strategy 
with five alternatives: (a) Using a fixed uniform group-scaling parameter $\sigma(\boldsymbol{\alpha})=\frac1N\boldsymbol{1}$ (\ie, no optimization), (b) minimizing group losses $\sigma(\boldsymbol{\alpha})^{\top}L(\theta)$, (c) MGDA that minimizes $\left\lVert \sigma(\boldsymbol{\alpha})^{\top}(\nabla_{\theta} L(\theta))_{\dagger}\right\rVert_2^2$, (d) GradNorm \citep{chen2018gradnorm}, (e) MoCo, the latest technique for MOO method \citep{fernando2023mitigating}, and (f) our method that minimizes $L_{\boldsymbol{\alpha}}$ in Eq.~(\ref{eqn:loss_alpha}).
Intriguingly, 
(b) leads to worse performance compared to (a) that uses a fixed value for $\boldsymbol{\alpha}$. 
We found that utilizing a learnable group-scaling parameter based solely on the weighted sum of group-wise losses resulted in worse performance in all metrics except \textsc{InDist} when compared with training without it.
The results in (c), (d), and (e) demonstrate that blindly applying an existing MOO method as-is with our grouping strategy
falls short of the desired level of unbiased performance during training on a biased dataset. This highlights the superiority of our method in scenarios involving multiple spurious correlations.

\begin{table*}[t]
\caption{
Ablation study on the grouping strategy on MultiCelebA in two biases setting: Grouping by bias attribute $\boldsymbol{b}$, grouping by both bias attribute and target class $(\boldsymbol{b}, t)$, and our strategy using the list of binary group labels $\boldsymbol{g}$. SUBG and GroupDRO with our grouping strategy are indicated by $\dagger$.
}
\begin{center}
\begin{small}
\begin{tabular}{lc|cccccc} %
\hline
Method & Group by & \textsc{InDist} &\textsc{GG} &\textsc{GC} &\textsc{CG} &\textsc{CC} & \textsc{Unbiased} \\ %
\hline
ERM  & - & \textbf{97.0}$_{\pm \text{0.2}}$ & \textbf{98.2}$_{\pm \text{0.7}}$ & \textbf{89.2}$_{\pm \text{2.6}}$ & 58.2$_{\pm \text{3.0}}$ & 19.0$_{\pm \text{1.8}}$ & 63.8$_{\pm \text{1.2}}$ \\
\hline
SUBG & $\boldsymbol{b},t$ & 80.3$_{\pm \text{1.1}}$ & 77.1$_{\pm \text{1.0}}$ & 78.4$_{\pm \text{0.7}}$ & 77.5$_{\pm \text{1.7}}$ & 78.0$_{\pm \text{1.2}}$ & 77.7$_{\pm \text{0.6}}$ \\
SUBG$^\dagger$ & $\boldsymbol{g}$ & 80.4$_{\pm \text{2.7}}$ & 78.5$_{\pm \text{4.3}}$ & 75.9$_{\pm \text{3.3}}$ & 71.9$_{\pm \text{2.2}}$ & 67.0$_{\pm \text{2.3}}$ & 73.3$_{\pm \text{1.7}}$ \\
\hline
GroupDRO & $\boldsymbol{b},t$ & 83.5$_{\pm \text{0.7}}$ & 81.2$_{\pm \text{1.0}}$ & 81.2$_{\pm \text{1.2}}$ & 76.7$_{\pm \text{1.5}}$ & 74.6$_{\pm \text{0.4}}$ & 78.4$_{\pm \text{0.7}}$ \\
GroupDRO$^\dagger$ & $\boldsymbol{g}$ & \underline{85.8}$_{\pm \text{1.5}}$ & \underline{83.1}$_{\pm \text{1.9}}$ & 79.5$_{\pm \text{2.4}}$ & \underline{80.7}$_{\pm \text{1.2}}$ & 71.8$_{\pm \text{1.1}}$ & 78.8$_{\pm \text{1.4}}$ \\ %
\hline
Ours & $\boldsymbol{b}$ & 79.2$_{\pm \text{0.7}}$ & 79.5$_{\pm \text{4.6}}$  & 79.8$_{\pm \text{3.5}}$ & 78.1$_{\pm \text{2.1}}$ & 77.0$_{\pm \text{1.6}}$ & 78.6$_{\pm \text{2.0}}$ \\
Ours & $\boldsymbol{b}, t$ & 78.5$_{\pm \text{5.5}}$ & 79.4$_{\pm \text{2.9}}$& 80.0$_{\pm \text{2.2}}$ & 79.0$_{\pm \text{1.9}}$ & \underline{78.5}$_{\pm \text{1.3}}$ & \underline{79.2}$_{\pm \text{1.4}}$ \\
\rowcolor{Gray}
Ours & $\boldsymbol{g}$ & 84.3$_{\pm \text{0.9}}$ & 82.4$_{\pm \text{0.9}}$ & \underline{85.1}$_{\pm \text{0.4}}$ & \textbf{81.7}$_{\pm \text{0.4}}$ & \textbf{82.6}$_{\pm \text{1.0}}$ & \textbf{82.9}$_{\pm \text{0.2}}$ \\ %
\hline
\end{tabular}
\end{small}
\end{center}
\label{ablation_group}
\vspace{-2mm}
\end{table*}

\begin{table}
    \caption{Impact of the update period $U$ of the group-scaling parameter on MultiCelebA in two biases setting.}
    \vspace{-2mm}
    \begin{center}
    \begin{small}
    \resizebox{\linewidth}{!}{
    \begin{tabular}{l|ccccc}
    \hline
    $U$ & \textsc{GG} & \textsc{GC} & \textsc{CG} & \textsc{CC} & \textsc{Unbiased}\\
    \hline 
    1 & 84.2$_{\pm \text{0.5}}$ & 86.0$_{\pm \text{0.5}}$ & 80.8$_{\pm \text{0.5}}$ & 80.8$_{\pm \text{0.5}}$ & 82.9$_{\pm \text{0.3}}$\\
    5 & 83.3$_{\pm \text{0.4}}$ & 85.8$_{\pm \text{0.7}}$ & 81.2$_{\pm \text{0.4}}$ & 81.7$_{\pm \text{0.1}}$ & 83.0$_{\pm \text{0.1}}$\\
    \rowcolor{Gray}
    10 & 82.4$_{\pm \text{0.6}}$ & 85.1$_{\pm \text{0.4}}$ & 81.7$_{\pm \text{0.3}}$ & 82.6$_{\pm \text{0.9}}$ & 82.9$_{\pm \text{0.2}}$  \\
    20 & 81.9$_{\pm \text{0.5}}$ & 84.9$_{\pm \text{0.5}}$ & 81.8$_{\pm \text{0.5}}$ & 83.0$_{\pm \text{0.9}}$ & 82.9$_{\pm \text{0.3}}$ \\
    30 & 79.3$_{\pm \text{1.3}}$ & 84.0$_{\pm \text{0.2}}$ & 82.6$_{\pm \text{0.3}}$ & 85.0$_{\pm \text{0.8}}$ & 82.7$_{\pm \text{0.2}}$\\
    \hline
    \end{tabular}
    }
    \end{small}
    \end{center}
    \vspace{-2mm}
    \label{ablation_m_table}
\end{table}

\noindent\textbf{Change of the group-scaling parameter over time.} 
We compare the trend of the group-scaling parameter in our method with that of GroupDRO~\citep{sagawa2019distributionally} on MultiCelebA in the two-bias setting, as illustrated in Figure~\ref{fig:alpha_graph}.
Our method shows an increasing trend for the weight of the \textsc{CC} group, while those of the other groups decrease during training.
This indicates that the model initially learns a shared representation that incorporates information from all the groups, but later focuses more on the minority group.
On the other hand, GroupDRO exhibits a decreasing weight trend for the minority groups (\textsc{CC} (L) and \textsc{CC} (H) in Figure~\ref{fig:alpha_graph}).
This trend occurs because the minority groups have lower training losses in the early stages of training, leading to lower weights in GroupDRO.
As a consequence, it tends to ignore minority groups and exacerbate the bias issue, resulting in inferior performance compared to the upweigthing baseline as shown in Table~\ref{tab:multiceleba_table}.

\noindent\textbf{Ablation study on the grouping strategy.}
To verify the contribution of our grouping strategy, 
we compare ours with two alternatives: grouping samples by the same bias attribute $\boldsymbol{b}$, and grouping those with the same pair of bias attribute $\boldsymbol{b}$ and target class $t$.
Figure~\ref{fig:ablation_grouping} demonstrates performance variations by different grouping policies.
To ensure a fair comparison, the test data are grouped by $\textbf{b}$ and $t$, which is the same as the conventional grouping strategy.
Figure~\ref{fig:ablation_grouping}(a) shows that the test accuracy gap between groups enlarges as training progresses when using the bias attribute grouping.
We conjecture that this problem arises from class imbalance within the groups categorized solely by bias attributes.
Specifically, the number of samples belonging to a target class that is spuriously correlated with the bias attribute becomes dominant, leading to an imbalanced representation of target classes within the group.
In Figure~\ref{fig:ablation_grouping}(b), we applied the commonly used strategy: grouping by both target classes and bias attributes. Compared with the conventional grouping, our method demonstrates a smaller performance gap between groups and higher the lowest group accuracy, as shown in Figure~\ref{fig:ablation_grouping}(c). 
Finally, we also report the performance metrics in Table~\ref{ablation_group}, 
which demonstrates
that our grouping strategy outperforms the others in all metrics.

\textbf{Applying our grouping strategy to GroupDRO and SUBG.} We also compare our method with GroupDRO and SUBG using the proposed grouping strategy. 
Results in Table~\ref{ablation_group} suggest that applying our grouping strategy alone to existing debiased training methods failed to achieve performance comparable to ours. This highlights the contribution of both our debiased training algorithm and grouping strategy to performance improvement.

\noindent\textbf{Impact of the update period $U$.}
We conducted experiments to examine how hyperparameter $U$ affects the performance of our method.
Table~\ref{ablation_m_table} reports the performance in \textsc{GG}, \textsc{GC}, \textsc{CG}, \textsc{CC} and \textsc{Unbiased} metrics on MultiCelebA using five different values of $U$.
To disregard the influence of the learning rate $\eta_2$, we adjusted the learning rate $\eta_2$ inversely proportional to the increase in the value of $U$. 
We found that the \textsc{Unbiased} remained consistent across all $U$ values we examined, which suggests that our algorithm is not sensitive to $U$. %

\section{Conclusion}
We have presented a novel debiased training algorithm that addresses the challenges posed by multiple biases in training data, inspired by multi-objective optimization (MOO). %
In addition, we have introduced a new real-image multi-bias benchmark, dubbed MultCelebA.
Our method surpassed existing algorithms for debaised training in both multi-bias and single-bias settings on six benchmarks in total.

\section*{Acknowledgement}
We express our gratitude to the reviewers for their thoughtful comments, which help improve our paper. This research was supported by the NRF grant and IITP grants funded by Ministry of Science and ICT (2021R1A2C3012728, %
RS-2022-II220290, %
RS-2019-II191906, %
IITP-2021-0-00739) %
and the NRF grant
funded by the Ministry of Education, Korea (2022R1A6A1A03052954). %

\section*{Impact Statement}
This paper presents work whose goal is to advance the field of Machine Learning. There are many potential societal consequences of our work, none which we feel must be specifically highlighted here.

\nocite{langley00}

\bibliography{egbib}
\bibliographystyle{icml2024}

\newpage
\appendix
\onecolumn

\begin{appendices}

\setcounter{figure}{0}
\setcounter{table}{0}
\renewcommand{\thefigure}{A\arabic{figure}}
\renewcommand{\thetable}{A\arabic{table}}
\renewcommand{\thesection}{A.\arabic{section}}
\renewcommand{\thesubsection}{A.\arabic{section}.\arabic{subsection}}

\setlength{\cftsecnumwidth}{8mm} %
\setlength{\cftsubsecnumwidth}{10mm} %

\section{Explaining spurious correlation in machine learning}
Spurious correlation refers to a relationship between variables that appears to be statistically significant but is actually caused by some other factor. For example, if most samples of class $a$ have an attribute $i$ and most samples of class $b$ have an attribute $j$, where $a \neq b$ and $i \neq j$, and neither attribute $i$ nor $j$ is not the actual cause of the target classes, then a trained model can rely on the bias attributes to classify most training samples. In this case, the bias attributes $i$ and $j$ can be considered as spuriously correlated with the target class, and each bias attribute is the bias-guiding attribute for its respective class.

\section{The construction process of MultiCelebA}\label{supp_section:diagonal}
In this section, we explain the construction of the two-bias setting of MultiCelebA, including the selection of the target class and bias types (Section~\ref{supp:two_bias_settings}). We then describe two additional evaluation settings: the three-bias setting and the four-bias setting (Section~\ref{supp:extra_eval_settings}).

\begin{minipage}{\linewidth}
    \centering
    \begin{minipage}[t]{0.39\linewidth}
    \centering
    \captionof{figure}{Unbiased accuracy for predicting each attribute}
    \vspace{-1mm}
    \includegraphics[width=0.75\linewidth]{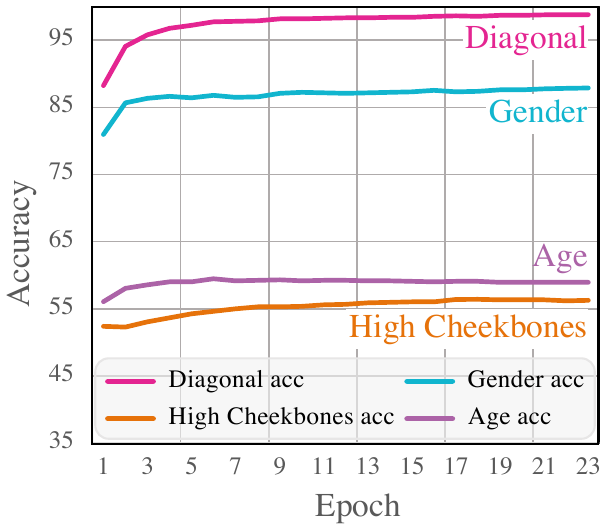}
    \label{fig:diagonal}
    \end{minipage}
    \hfill
    \begin{minipage}[t]{0.6\linewidth}
    \centering
        \captionof{table}{Configuration of MultiCelebA in two biases setting}
        \label{tab:configure_two_biases}
        \vspace{3mm}
        \begin{small}
        \begin{tabular}{c|cc}
             \multirow{2}{*}{Group} & \multirow{2}{*}{\{Target class, Bias type 1, Bias type 2\}} & \# of training\\ 
             & & samples\\
             \hline
             \multirow{2}{*}{\textsc{GG}} & \{\texttt{High Cheekbones}, \texttt{Female}, \texttt{Young}\} & 44582\\
             &\{\texttt{Low Cheekbones}, \texttt{Male}, \texttt{Old}\} & 16220 \\
             \hline
             \multirow{2}{*}{\textsc{GC}} & \{\texttt{High Cheekbones}, \texttt{Female}, \texttt{Old}\} & 2200\\
             & \{\texttt{Low Cheekbones}, \texttt{Male}, \texttt{Young}\} & 800\\
             \hline
             \multirow{2}{*}{\textsc{CG}} & \{\texttt{High Cheekbones}, \texttt{Male}, \texttt{Young}\} & 2200\\
             & \{\texttt{Low Cheekbones}, \texttt{Female}, \texttt{Old}\} & 800\\
             \hline
             \multirow{2}{*}{\textsc{CC}} & \{\texttt{High Cheekbones}, \texttt{Male}, \texttt{Old}\} & 110\\
             & \{\texttt{Low Cheekbones}, \texttt{Female}, \texttt{Young}\} & 40\\
             \hline
        \end{tabular}
        \end{small}
    \end{minipage}
\end{minipage}

\subsection{Two-bias setting of MultiCelebA}\label{supp:two_bias_settings}
We first selected \texttt{gender} and \texttt{age} as bias types among the 40 attributes of CelebA. We then chose \texttt{high-cheekbones} as the target class and verified whether the target class and both bias types exhibit spurious correlations and invoke shortcut learning for ERM.

\citet{scimeca2021shortcut} examined how deep neural networks exhibit a preference for attributes based on their ease of learning.
Following \citet{scimeca2021shortcut}, we assessed the preference of ResNet18 for the target class (\texttt{high-cheekbones}) and biases (\texttt{gender} and \texttt{age}) by evaluating a model trained on diagonal set (\textsc{GG} group in the main paper), where all samples are spuriously correlated with all biases, as shown in Figure~\ref{fig:diagonal}.
Each line on Figure~\ref{fig:diagonal} represents unbiased accuracy of a testing attribute, which we used to evaluate the model's ability to predict each testing attribute. 
ResNet18 exhibited higher unbiased accuracy for \texttt{gender} and \texttt{age} compared to that of \texttt{high-cheekbones}, indicating that the model tends to exploit \texttt{gender} and \texttt{age} as shortcuts when learning \texttt{high-cheekbones} classification task on MultiCelebA.
With \texttt{high-cheekbones}, \texttt{gender}, and \texttt{age} labels, we subsampled the training set of CelebA to simulate challenging scenarios where training data are extremely biased. 
The configurations of MultiCelebA in the two-bias setting are shown in Table~\ref{tab:configure_two_biases}.

\begin{table}[h]
    \caption{Performance in \textsc{CCCC} and \textsc{Unbiased} (\%) on MultiCelebA in four biases for evaluation two biases for training setting.}
    \centering
    \begin{small}
    \begin{tabular}{lcc}
        \hline
        Method & CCCC & \textsc{Unbiased}\\
        \hline
        ERM & 6.0$_{\pm \text{0.9}}$ & 59.0$_{\pm \text{0.7}}$\\
        GroupDRO & 29.6$_{\pm \text{3.7}}$ & 62.3$_{\pm \text{0.9}}$\\
        Ours & 43.1$_{\pm \text{1.5}}$ & 65.8$_{\pm \text{0.2}}$\\
        \hline
    \end{tabular}
    \end{small}
    \label{tab:4types}
\end{table}

\subsection{three-bias and four-bias settings of MultiCelebA}\label{supp:extra_eval_settings}
Next, we extend our MultiCelebA dataset with two bias types by adding extra bias types for further evaluation settings. 
To this end, we explain which attributes can be considered as bias types. For evaluating debiased training algorithms, the training set should be designed in such a way that ERM exploits undesirable shortcuts stemming from spurious correlations between the target labels and predefined bias types (\ie, the ERM solution trained on the set is biased). To this end, the selected bias types have to hold the two conditions below:
\vspace{-2mm}
\begin{itemize}
    \setlength\itemsep{-0.5mm}
    \item The bias types are all spuriously correlated to the target class.
    \item Not every sample in the training set has the same labels for a pair of the bias types.
\end{itemize}
\vspace{-2mm}
The first condition is trivial, and the second is required to reject bias-type candidates redundant to those previously chosen.

Among the 37 attributes of CelebA, we identified \texttt{mouth slightly open} as a third bias type, and then we chose \texttt{smiling} as a fourth bias type, both of which satisfy the two conditions. 
We demonstrated the superior results on the three-bias setting and the four-bias setting, as shown in Table~\ref{tab:3types_table} and Table~\ref{tab:4types}, respectively.

\subsection{Considerations on exceeding four bias types}\label{supp_subsubsection:more}
We empirically found that it is difficult to identify more than five attributes (\ie, bias types) that satisfy the two conditions at once in existing natural image datasets, even for CelebA with 40 attributes. Furthermore, some groups often become empty sets when the number of bias types increases; this is a critical issue particularly for test data as some groups with zero cardinality are never used for evaluation.

\begin{figure}[h]
\begin{minipage}[t]{0.49\linewidth}
\centering
\includegraphics[width=\linewidth]{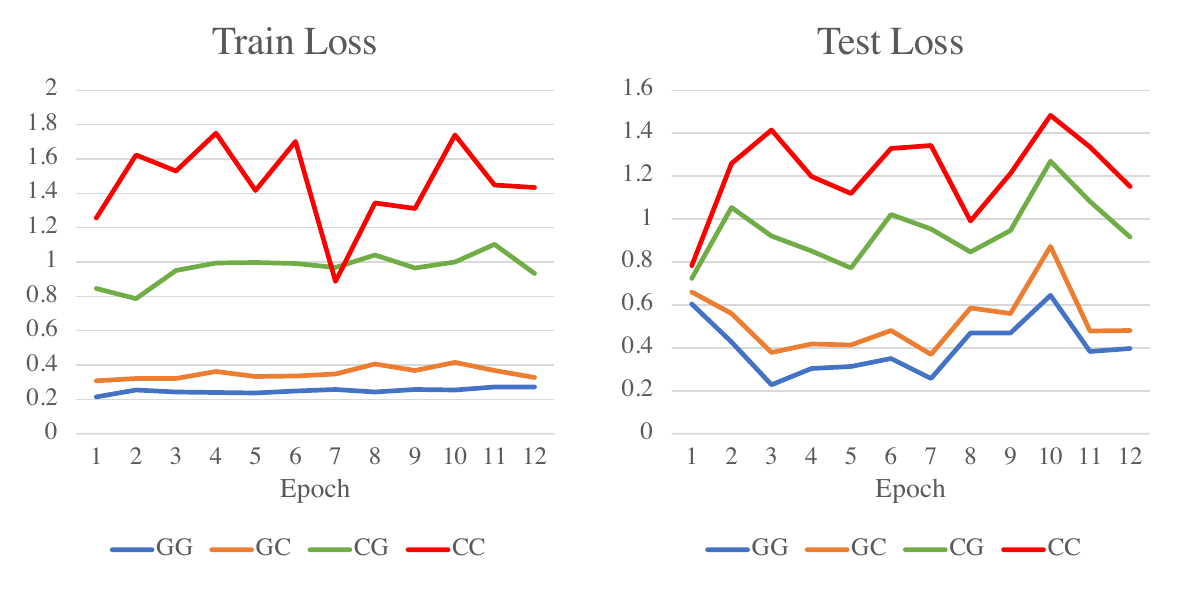}
\caption*{(a) Train with ERM}
\label{fig:loss_ERM}
\end{minipage}
\hfill
\begin{minipage}[t]{0.49\linewidth}
\centering
\includegraphics[width=\linewidth]{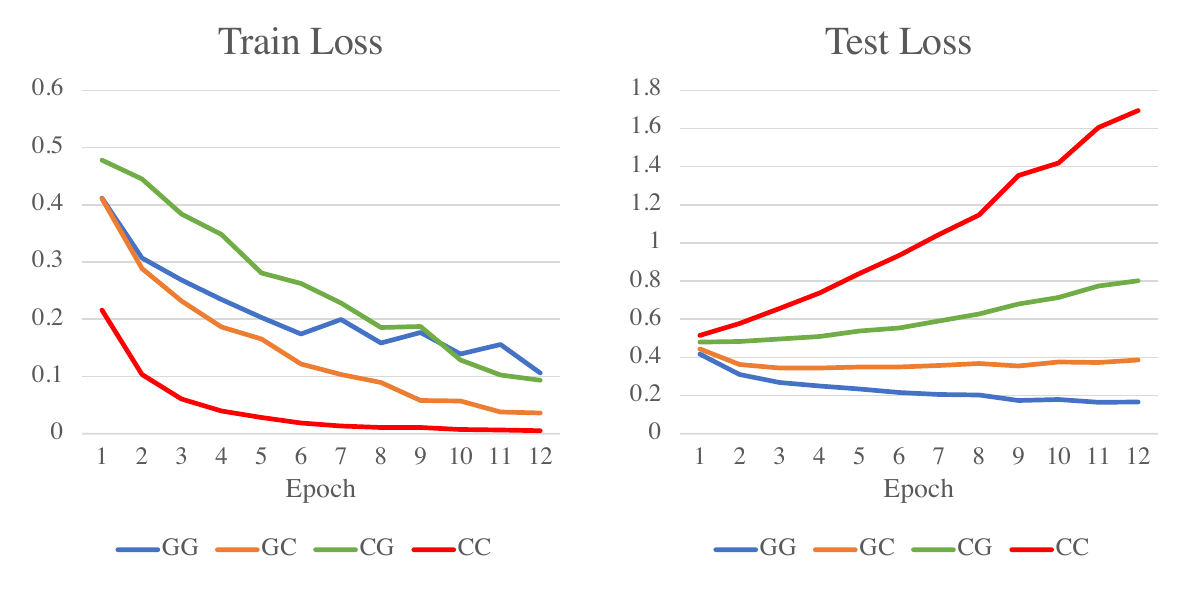}
\caption*{(b) Train with the averaged group losses}
\label{fig:loss_avg_group}
\end{minipage}
\caption{Group losses of (a) model with ERM (b) model with the averaged group losses.}
\label{fig:loss}
\end{figure}

\section{Empirical analysis of the objective for optimizing the group-scaling parameter}\label{supp_section:loss}
When training a model by ERM, the training loss for the small group is larger than that for the large group, and a similar trend is observed in the test loss, as shown in Figure~\ref{fig:loss}(a). 
Thus, increasing the weight of groups with a larger training loss can be beneficial in giving more weight to the minority group.

However, when we compute the objective by averaging group-wise losses, the gap between training loss and test loss for each group varies depending on group size, as shown in Figure~\ref{fig:loss}(b). This phenomenon arises because smaller groups are more susceptible to memorization effects. 

To mitigate the gap between training loss and test loss resulting from the memorization effect, \citet{sagawa2019distributionally} proposed the use of strong regularization on model parameters and an increase in the weight of group with large training loss. 
This approach of increasing the weight of groups or samples with large train loss has evolved into a standard practice within debiased training methods.

However, in scenarios involving multiple biases, a trained model easily overfits minor groups (\eg, the CC group) that have a small number of samples, leading to decreased loss scales for such groups and consequently neglecting them in training, resulting in a biased model.
Applying a strong $\ell_2$ regularizer to model parameters was successful in the single bias settings, but we empirically found that it does not work as desired in multi-bias settings. This is particularly due to minor groups, \eg, the CC group in the two-bias setting, having an extremely small number of samples.
That is the underlying cause of the inferior performance of GroupDRO in multiple biases settings compared to Upweighting.

In contrast, our algorithm optimizes the group-scaling parameter based on MOO. In a nutshell, it increases the weights of groups with low loss magnitudes. Since minor groups usually exhibit low loss scales as a model easily overfits them due to their small cardinality, our algorithm emphasizes the impact of minor groups in training.

\section{Implementation details}\label{supp_section:details}
\subsection{Datasets}
To evaluate our framework, we consider three multi-bias datasets, \ie, MultiCelebA, Multi-Color MNIST, and UrbanCars and three single-bias datasets, \ie, Waterbirds, CelebA, and BFFHQ. In what follows, we provide details of each dataset.

\textbf{MultiCelebA.}
First, we mainly consider MultiCelebA in two biases setting as the dataset to evaluate debiased training algorithms. As introduced in Section~\ref{MultiCelebA}, this dataset requires training a model to predict whether if a given face image has \texttt{high-cheekbones} or not. Each image is additionally annotated with \texttt{gender} and \texttt{age} attributes which are spuriously correlated with the target \texttt{high-cheekbones}.
For MultiCelebA in three biases setting, each image is annotated with \texttt{gender}, \texttt{age}, and \texttt{mouth slightly open} attributes which are spuriously correlated with the target \texttt{high-cheekbones}.

\textbf{UrbanCars.} UrbanCars~\citep{li2023whac} is a dataset created by synthesizing \texttt{background}, \texttt{co-occurring object}, and \texttt{car} to generate multi-biased images. Its task involves classifying whether an image contains \texttt{urbancars} or not.

\textbf{Multi-Color MNIST.} We consider Multi-Color MNIST dataset proposed by \citet{li2022discover}. Its task is to predict the digit number from an image. The digit numbers are spuriously correlated with left and right background colors, coined \texttt{left-color} and \texttt{right-color}, respectively. As proposed by Li~\etal, we set the proportion of bias-guiding attributes to be 99\% and 95\% for \texttt{left-color} and \texttt{right-color}, respectively.

\textbf{Waterbirds.} Waterbirds~\citep{sagawa2019distributionally} is a single-bias dataset consisting of bird images. Given an image, the target is \texttt{bird-type}, \ie, whether if the bird is ``landbird'' or a ``waterbird.'' The biased attribute is \texttt{background-type}, \ie, whether if the image contains ``land'' or ``water.'' The proportion of biased attribute is set to 95\%.

\textbf{CelebA.} CelebA~\citep{CelebA} is a face recognition dataset where each sample is labeled with 40 attributes. Following the previous settings~\citep{sagawa2019distributionally,yao2022improving}, we use \texttt{HairColor} as the target and \texttt{gender} as the bias attribute.  

\textbf{BFFHQ.} BFFHQ~\citep{LDR} is a real-world face image dataset curated from FFHQ. 
Its task is to predict the age from an image. the age is spuriously correlated with gender attributes. The proportion of bias-guiding attributes is 99.5\%.

\subsection{Baselines}
We extensively compare our algorithm against the existing debiased training algorithms. In particular, one can categorize a baseline by whether it explicitly uses the supervision on biased attributes, \ie, bias labels, or not. 
To this end, compare our method with nine training algorithms, consisting of five that do not use the bias label and six that do. 
Algorithms that do not require using the bias label are as follows:
(1) training with vanilla ERM, 
(2) LfF~\citep{LfF} employs a reweighting scheme where samples that are more likely to be misclassified by a biased model are assigned higher weights, 
(3) JTT~\citep{JTT} retrains a model using different weights for each group, where the groups are categorized as either bias-guiding or bias-conflicting based on an ERM model,
(4) EIIL~\citep{creager2021environment} conducts domain-invariant learning, (5) PGI~\citep{ahmed2020systematic} matches the class-conditional distribution of groups by introducing predictive group invariance,
and (6) DebiAN~\citep{li2022discover} utilizes a pair of alternate networks to discover and mitigate unknown biases sequentially. 
We consider debiased training methods using bias attribute labels as follows: 
(1) Upsampling assigns higher sampling probability to minority groups, 
(2) Upweighting assigns scales the sample-wise loss to be higher for minority groups;
$\text{group weight} = (\#\text{ of training samples})/(\#\text{ of group samples})$,
(3) GroupDRO~\citep{sagawa2019distributionally} computes group-scaling weights using group-wise training loss to upweight the worst-case group samples. 
(4) SUBG~\citep{sagawa2020investigation} proposes a group-balanced sampling scheme by undersampling the majority groups. 
(5) LISA~\citep{yao2022improving} performs group mixing (mixup) augmentation to learn from both intra- and inter-group information. 
(6) DFR~\citep{kirichenko2022last} retrains the last layer of an ERM model using a balanced set obtained through undersampling.

\begin{table}[h]
\begin{center}
\begin{small}
\caption{
The search spaces of hyperparameters. %
}
\begin{tabular}{cc}
\hline
 Hyperparameter & Search space \\
\hline
\multirow{3}{*}{Learning rate $\eta_1$, $\eta_2$} & \{$\expnum{5}{4}$, $\expnum{2}{4}$, $\expnum{1}{4}$,\\
&$\;\;\expnum{5}{3}$, $\expnum{2}{3}$, $\expnum{1}{3}$, \\
&$\;\;\;\expnum{5}{2}$, $\expnum{2}{2}$, $\expnum{1}{2}$\}\\
Weight decay & \{0, $\expnum{1}{4}$, $\expnum{1}{2}$, $\expnum{1}{1}$, 1\}\\
Update frequency $U$& \{1, 5, 10, 50\} \\
\hline
\end{tabular}
\label{tab:hyperparameter_space}
\end{small}
\end{center}
\end{table}

\begin{table}[h]
\begin{center}
\begin{small}
\caption{
Hyperparameters of our method.%
}
\begin{tabular}{l|cccccc}
\hline
& MultiCelebA & Multi-Color MNIST & UrbanCars & Waterbirds & CelebA & BFFHQ\\
\hline
Batch size & 512 & 512 & 128 & 128 & 128 & 64 \\
Learning rate $\eta_1$ & $\expnum{2}{4}$ & $\expnum{2}{2}$ & $\expnum{1}{2}$ & $\expnum{1}{3}$ & $\expnum{2}{3}$ & $\expnum{2}{3}$\\ 
Learning rate $\eta_2$ & $\expnum{1}{2}$ & $\expnum{2}{3}$ & $\expnum{1}{3}$ & $\expnum{1}{3}$ & $\expnum{1}{4}$ & $\expnum{5}{4}$\\
Update frequency $U$ & 10 & 50 & 10 & 5 & 1 & 1\\
Optimizer & SGD & Adam & SGD & SGD & Adam & Adam\\

\hline
\end{tabular}
\label{tab:hyperparameters}
\end{small}
\end{center}
\end{table}

\subsection{Hyperparameters}
We tune all hyperparameters, as well as early stopping, based on highest \textsc{Worst} for MultiCelebA, UrbanCars, Waterbirds, and CelebA on validation set, except for ERM. For Multi-Color MNIST and BFFHQ, we tune hyperparemters based on highest \textsc{Unbiased} on test set, following the previous work~\citep{li2022discover, LDR}. 
We use a single GPU (RTX 3090) for training.
Following the previous work~\citep{LDR, hwang2022selecmix}, we conduct experiments on BFFHQ using ResNet18 with random initialization as the neural network architecture. 
For remaining datasets, we initialized the model with parameters pretrained on ImageNet.
The hyperparameter search spaces used in all experiments conducted in this paper are summarized in Table~\ref{tab:hyperparameter_space}. The selected hyperparameters for our method are represented in Table~\ref{tab:hyperparameters}.
Furthermore, the search space for the upweight value $\lambda_{up}$ in JTT is \{5, 10, 20, 30, 40, 50, 100\}.
JTT~\citep{JTT} and DFR~\citep{kirichenko2022last} utilize the ERM model as a pseudo labeler and frozen backbone network, respectively. 
We used the ERM model as reported in the literature for our implementation of these methods.
We did not use a learning rate scheduler in any of the experiments.

Given that the proportion of samples from minority groups can impact the performance of debiased training, we trained DFR exclusively on the training set to ensure a fair comparison, which is denoted as DFR$_{tr}^{tr}$.

\subsection{Training existing methods on multi-bias setting}
When training a model using SUBG~\citep{sagawa2020investigation}, GroupDRO~\citep{sagawa2019distributionally} and DFR~\citep{kirichenko2022last}, we grouped the training set based on the same pair of bias attribute $\boldsymbol{b}$ and target class $t$ and followed the approach outlined in the original paper.

LISA~\citep{yao2022improving} adopts the two kinds of selective augmentation strategies, Intra-label LISA and Intra-domain LISA. In the multi-bias setting, Intra-label LISA (LISA-L) interpolates samples with the same target label but different all bias labels ($t^{(m)}=t^{(m')}$, $b_d^{(m)} \neq b_d^{(m')}\;\forall d$).
Intra-domain LISA (LISA-D) interpolates samples with the same bias labels but different target label ($t^{(m)}\neq t^{(m')}$, $\boldsymbol{b}^{(m)}=\boldsymbol{b}^{(m')}$).

When training a model using biased training methods that do not require bias labels, such as LfF~\citep{LfF}, JTT~\citep{JTT}, and DebiAN~\citep{li2022discover}, we followed the approach outlined in the original paper without modification, regardless of the number of bias types presented in the dataset.

\subsection{Evaluation metrics}
We consider various metrics to evaluate whether if the trained model is biased towards a certain group in the dataset. We remark that no metric is universally preferred over others, \eg, worst-group and average-group accuracy reflects different aspects of a debiased training algorithm.

For the multi-bias datasets, we evaluate algorithms using the average accuracy for each of the four groups categorized by the guiding or conflicting nature of the biases: \{\textsc{GG}, \textsc{GC}, \textsc{CG}, \textsc{CC}\}. Here, \textsc{G} and \textsc{C} describes whether a group contains bias-guiding or bias-conflicting samples for each bias type, respectively. For example, \text{GC} group for MultiCelebA is an intersection of bias-guiding samples with respect to the first bias type, \ie, \texttt{gender}, and bias-conflicting samples with respect to the second bias type, \ie, \texttt{age}. We also report the average of these four metrics, denoted as \textsc{Unbiased}.
Conceptually, the GG metric can be high regardless of whether a model is biased or not. However, the CC metric can be high only when a model is debiased from all spurious correlations. Similarly, the GC accuracy can be high only when a model is debiased from the spurious correlation of the second bias type. Meanwhile, the InDist accuracy measures the average accuracy on biased data. A biased model will achieve high scores in InDist and GG metrics, but low scores in Unbiased and CC metrics. This also means that huge performance variations among GG, GC, CG, and CC groups suggest model bias to spurious correlations (as shown in the scores of ERM). The effectiveness of debiased training algorithms can be assessed by examining the Unbiased accuracy first (higher is better) and in turn the CC/Worst accuracy (higher is better); the GC, CG, and GG accuracies also have to be sufficiently high compared to the CC accuracy.

Next, for the single-bias datasets, the minimum group average accuracy is reported as \textsc{Worst}, and the weighted average accuracy with weights corresponding to the relative proportion of each group in the training set as \textsc{InDist} (in-distribution) following \citet{sagawa2019distributionally}.

In calculating the {\textsc{GG}, \textsc{GC}, \textsc{CG}, \textsc{CC}} accuracies on the MultiCelebA dataset, we excluded the impact of class imbalance within each group by first computing the mean accuracy for each class within the group, and then taking the average of the class accuracies to obtain the group accuracy.

\subsection{Interpretation of the results on MultiCelebA}
In Table 1, we analyzed whether a model is biased toward the two bias types based on the difference between {\textsc{GG}, \textsc{GC}, \textsc{CG}, \textsc{CC}}, while also evaluating the \textsc{Unbiased} accuracy.
Let \textsc{G*} denote the combination of \textsc{GG} and \textsc{GC}, and similarly for \textsc{C*} and others.
A model is biased toward \texttt{gender} attribute if there is a significant difference between the \textsc{G*} and \textsc{C*} combinations, whereas a significant difference between the \textsc{*G} and \textsc{*C} combinations indicates bias toward \texttt{age} attribute.

\begin{minipage}[h]{\linewidth}
    \begin{minipage}[t]{0.47\linewidth}
    \captionof{table}{The number of groups for training on Multi-Color MNIST, UrbanCars, MultiCelebA, Waterbirds, CelebA,  and BFFHQ.}
    \centering
    \resizebox{0.7\linewidth}{!}{
    \begin{tabular}{lcc}
    \hline
    Benchmark & GroupDRO & Ours\\
    \hline
    Multi-Color MNIST & 40 & 4 \\
    UrbanCars & 8 & 4 \\
    MultiCelebA & 8 & 4 \\
    Waterbirds & 4 & 2 \\
    CelebA & 4 & 2 \\
    BFFHQ & 4 & 2 \\
    \hline
    \end{tabular}
    }
    \label{tab:the_number_of_groups}
  \end{minipage}
  \hfill
  \begin{minipage}[t]{0.47\linewidth}
    \centering
    \captionof{figure}{Local loss curvature of the loss landscape of model parameter.}
    \includegraphics[width=0.7\linewidth]{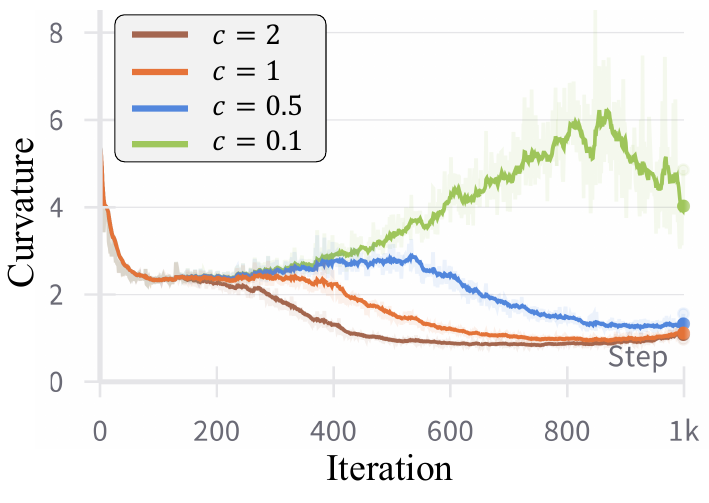}
    \label{fig:supp_curvature}
    \end{minipage}
    \vspace{-3mm}
\end{minipage}

\section{Impact of the loss function on local curvature}\label{supp_section:curvature}
According to \citet{li2021robust}, the second term in Eq.~(\ref{eqn:loss_alpha}), $\left\lVert \sigma(\boldsymbol{\alpha})^{\top}(\nabla_{\theta} L(\theta))\right\rVert_2^2$, serves as an approximation for the local curvature of the loss landscape associated with the model parameter $\theta$. Although this term is minimized by updating $\alpha$, the local curvature of loss landscape of model parameter is reduced. To verify this, we conducted an ablation study by adjusting the relative weight of the second term in Eq.~(\ref{eqn:loss_alpha}) using constant $c$. The objective formula for this experiment is presented as:
\begin{align}
    \hat{L}_{\boldsymbol{\alpha}} = \sigma(\boldsymbol{\alpha})^{\top}L(\theta) +c\lambda\left\lVert \sigma(\boldsymbol{\alpha})^{\top}(\nabla_{\theta} L(\theta))_{\dagger}\right\rVert_2^2.
\end{align}
We updated $\boldsymbol{\alpha}$ and $\lambda$ by minimizing $\hat{L}_{\boldsymbol{\alpha}}$ and $\theta$ by minimizing Eq.~(\ref{eqn:loss_theta}).
Figure~\ref{fig:supp_curvature} demonstrates how the loss curvature evolves over training iterations. We observed that as the value of $c$ decreases, there is a corresponding increase in loss curvature. Hence, minimizing the second term in Eq.~(\ref{eqn:loss_alpha}) contributes to improving model generalization.

\section{Comparison of the number of groups in previous methods and our method}
We compared the number of groups for training with GroupDRO on all the benchmarks we used. For all experiments in our paper, the same annotations have been used across ours and the other methods using bias labels, \ie, upsampling, upweighting, SUBG, GroupDRO, LISA, and DFR. 
For example, on multiple bias settings with two bias types, all the aforementioned methods including ours utilize labels for two bias types for group division.
As shown in Table~\ref{tab:the_number_of_groups}, our method defines a smaller number of groups compared with GroupDRO for debiased training. The number of groups of GroupDRO, denoted as $N_{\text{GroupDRO}}$, and that of ours, denoted as $N_{\text{Ours}}$, are calculated as follows:
\begin{align}
N_{\text{GroupDRO}}&=C\times(\# \text{of attributes in bias type 1})\times…\times(\# \text{of attributes in bias type D})\ge 2^D\times C, \\
N_{\text{Ours}}&=2 \times 2 \times…\times 2=2^D,
\end{align}
where $C$ is the number of classes, $D$ is the number of bias types, and $2^D\times C$ is the lower bound of $N_{\text{GroupDRO}}$. Since the number of classes $C$ is greater than 1, $N_{\text{Ours}}$ is always smaller than $N_{\text{GroupDRO}}$. The number of groups for each dataset is presented in Table~\ref{tab:the_number_of_groups}.

\section{Analysis of the limitation of CivilComments as a multi-bias setting}
CivilComments has been used to benchmark debiased training algorithms in a single spurious correlation setting~\cite{borkan2019nuanced, koh2021wilds}. Its target task is to classify an online comment into toxic or non-toxic, and the class label is spuriously correlated with certain demographic identities (e.g., male, female, White, Black, LGBTQ, Muslim, Christian, and other religions) mentioned in the comment.

To investigate whether CivilComments involves multiple spurious correlations, we first categorized the demographic labels into 3 types: Gender: \{male, female, LGBTQ\}, Race: \{White, Black\}, Religions: \{Muslim, Christian, and other religions\}.
We then examined if these 3 types are spuriously correlated with the target class (\ie, toxic). If most samples of class $a$ have a bias attribute $i$ and most samples of class $b$ have a bias attribute $j$ ($a \neq b$ and $i \neq j$), then a trained model can rely on the bias attributes to classify most training samples, and the bias attributes $i$ and $j$ can be considered as spuriously correlated with the target class in this case. Based on this notion, we investigated the type-wise data population of the dataset as shown in Table~\ref{civilcomments_analysis}, and found that none of the three bias types are spuriously correlated with the toxic class.
\begin{table}[h]
\begin{center}
\begin{small}
\caption{Training data population of non-toxic and toxic comments based on identity presence across gender, race, and religion.}
\label{civilcomments_analysis}
\begin{tabular}{l|cc|cc|cc}
\hline
& \multicolumn{2}{c|}{Gender} & \multicolumn{2}{c|}{Race} & \multicolumn{2}{c}{Religion}\\
 & no identities & has identities & no identities & has identities & no identities & has identities \\
\hline
non-toxic & 188585 (70\%) & 49938 (19\%) & 202071 (75\%) & 36452 (14\%) & 222348 (83\%) & 16175 (6\%) \\
toxic & 21207 (8\%) & 9308 (3\%) & 24852 (9\%) & 5663 (2\%) & 24000 (9\%) & 6515 (2\%) \\
\hline
\end{tabular}
\end{small}
\end{center}
\end{table}

\section{Computational complexity}
To demonstrate the scalability of our algorithm, we show that the overall computational complexity of our algorithm grows slower than the number of groups $2^D$ by the proof below.

\textbf{Preliminary:}

There are six factors contributing to the total computational complexity of our debiased training algorithm, as enumerated below:
\vspace{-2mm}
\begin{itemize}
    \setlength\itemsep{-1mm}
    \item $D$: the number of bias types (the number of groups is then $2^D$)
    \item $U$: the update period of $\boldsymbol{\alpha}$ (e.g., $U=10$ means $\boldsymbol{\alpha}$ is updated every 10 iterations.)
    \item $a$: the computational complexity for forward process per epoch
    \item $b$: the computational complexity for backward process per epoch
    \item $c$: the computational complexity for model parameter update per epoch
    \item $d$: the computational complexity for $\boldsymbol{\alpha}$ update per epoch
    \item $e$: the computational complexity for $\lambda$ update per epoch
\end{itemize}
\vspace{-2mm}
The overall computation complexity of our algorithm with $D$ bias types for each epoch is then denoted and defined by $C_D := a+b+c + 2^D\times b/U +d/U +e/U$.

\textbf{Proposition:} The overall complexity ($C_D$) grows slower than the number of groups ($2^D$).

\textbf{Proof by Contradiction:} We first assume a negation of the proposition: ``The overall complexity increases at least linearly with the number of groups." 

Under this assumption, $C_{D+1}\ge 2 \cdot C_D$, where $D\geq1$. 

Then, regarding $C_D := a+b+c + 2^D\times b/U +d/U +e/U$,

$C_{D+1}\ge 2 \cdot C_D$

$\Leftrightarrow a+b+c + 2^{D+1}\times b/U +d/U +e/U\ge 2a+2b+2c + 2^{D+1}\times b/U +2d/U +2e/U$

$\Leftrightarrow 0\ge a+b+c+d/U+e/U$,

which is a contradiction since the right-hand side is always greater than 0. Therefore, the assumption is false and the proposition holds.

The proposition suggests that \textit{the total time complexity of training in our algorithm grows slower than the number of groups} (\ie, $2^D$); a simple analysis reveals that $O(C_D)=2^D$ \textbf{when} $D$ \textbf{goes to the infinity}. 

To empirically verify this conclusion, we further increased the number of bias types of MultiCelebA up to four and estimated the wall-clock time of our training algorithm versus the number of bias types. As demonstrated in Table~\ref{tab:time_complexity}, the wall-clock time does not increase exponentially, even the number of groups $N$ is exponentially increased in the number of bias types $D$. 
\begin{table}[h]
    \caption{Comparison of time complexity according to the number of bias types}
    \label{tab:time_complexity}
    \centering
    \begin{small}
    \begin{tabular}{ccccc}
        \hline
        \# of & \# of & Training time & Relative training time & Relative training time\\
        bias types (D) & groups (N) & per 1 epoch  & (compare to D=2) & (compare to D=3)\\
        \hline
        2 & 4 & 67s	 & - & - \\
        3 & 8 & 82s	 & 1.22 times & - \\
        4 & 16 & 113s & 1.69 times & 1.37 times \\
        \hline
    \end{tabular}
    \end{small}
\end{table}

\end{appendices}

\end{document}